\pdfoutput=1

\documentclass[11pt]{article}

\usepackage{acl}
\usepackage{times}
\usepackage{latexsym}
\usepackage{amsmath}
\usepackage{graphicx}
\usepackage{multirow}
\usepackage{multicol}
\usepackage[hang,flushmargin]{footmisc}
\usepackage{algorithm}
\usepackage{algpseudocode}
\usepackage{courier}
\usepackage{array}
\usepackage{soul}
\usepackage{subcaption}
\usepackage{tcolorbox}

\usepackage{booktabs} 

\usepackage{calc}
\usepackage{graphicx}

\usepackage[T1]{fontenc}

\usepackage[utf8]{inputenc}

\usepackage{microtype}

\usepackage{inconsolata}
\usepackage{url}
\usepackage{hyperref,xcolor}

\definecolor{cornflowerblue}{rgb}{0.39, 0.58, 0.93}
\definecolor{babypink}{rgb}{0.99, 0.26, 0.76}

\newcommand{\hldarkblue}[1]{\begingroup\setlength{\fboxsep}{1.2pt}\colorbox{cornflowerblue!125}{\textcolor{white}{#1}}\endgroup}

\newcommand{\hlmedblue}[1]{\begingroup\setlength{\fboxsep}{1.2pt}\colorbox{cornflowerblue!75}{\textcolor{white}{#1}}\endgroup}

\newcommand{\hllightblue}[1]{\begingroup\setlength{\fboxsep}{1.2pt}\colorbox{cornflowerblue!20}{\textcolor{black}{#1}}\endgroup}

\title{\textit{An image speaks a thousand words, but can everyone listen?} \\On image transcreation for cultural relevance}

\author{Simran Khanuja \quad Sathyanarayanan Ramamoorthy   
\\  \vspace{1mm}{\centering {\bf Yueqi Song \quad  Graham Neubig}} \\
Carnegie Mellon University \\
\texttt{\{skhanuja, sramamoo, yueqis, gneubig\}@andrew.cmu.edu}}

\begin{document}
\newcommand{\refalg}[1]{Algorithm \ref{#1}}
\newcommand{\refeqn}[1]{Equation \ref{#1}}
\newcommand{\reffig}[1]{Figure \ref{#1}}
\newcommand{\reftbl}[1]{Table \ref{#1}}
\newcommand{\refsec}[1]{Section \ref{#1}}
\newcommand{\refapp}[1]{Appendix \ref{#1}}
\definecolor{coralpink}{rgb}{0.97, 0.51, 0.47}
\definecolor{babyblueeyes}{rgb}{0.63, 0.79, 0.95}

\newcommand{\todo}[1]{\textcolor{red}{[[ #1 ]]}\typeout{#1}}

\newcommand{\m}[1]{\mathcal{#1}}
\newcommand{\bmm}[1]{\bm{\mathcal{#1}}}
\newcommand{\real}[1]{\mathbb{R}^{#1}}
\newcommand{\method}{\textsc{Method}}

\newtheorem{theorem}{Theorem}[section]
\newtheorem{claim}[theorem]{Claim}

\newcommand{\argmax}{arg\,max}
\newcommand\norm[1]{\left\lVert#1\right\rVert}

\newcommand{\note}[1]{\textcolor{blue}{#1}}

\newcommand*{\Scale}[2][4]{\scalebox{#1}{$#2$}}%
\newcommand*{\Resize}[2]{\resizebox{#1}{!}{$#2$}}%

\newcommand{\SK}[1]{\textcolor{cyan} {[\textsc{sk}: #1]}}

\maketitle

\begin{abstract}
   Given the rise of multimedia content, human translators increasingly focus on culturally adapting not only words but also other modalities such as images to convey the same meaning. While several applications stand to benefit from this, machine translation systems remain confined to dealing with language in speech and text. In this work, we introduce a new task of translating \emph{images} to make them culturally relevant. First, we build three pipelines comprising state-of-the-art generative models to do the task. Next, we build a two-part evaluation dataset -- (i) \emph{concept}: comprising 600 images that are cross-culturally coherent, focusing on a single concept per image; and (ii) \emph{application}: comprising 100 images curated from real-world applications. We conduct a multi-faceted human evaluation of translated images to assess for cultural relevance and meaning preservation. We find that as of today, image-editing models fail at this task, but can be improved by leveraging LLMs and retrievers in the loop. Best pipelines can only translate 5\% of images for some countries in the easier \emph{concept} dataset and no translation is successful for some countries in the \emph{application} dataset, highlighting the challenging nature of the task. Our code and data is released here.\footnote{\url{https://github.com/simran-khanuja/image-transcreation}} 
\end{abstract}
\section{Introduction}

\begin{quote}
\emph{We shall try... to make not word-for-word but sense-for-sense translations.}
\end{quote}

\vspace{-0.55cm} 

\begin{flushright}
- \citet{jerome_pammachius}
\end{flushright}

\begin{figure*}[h]
    \centering
    \includegraphics[width=0.98\textwidth]{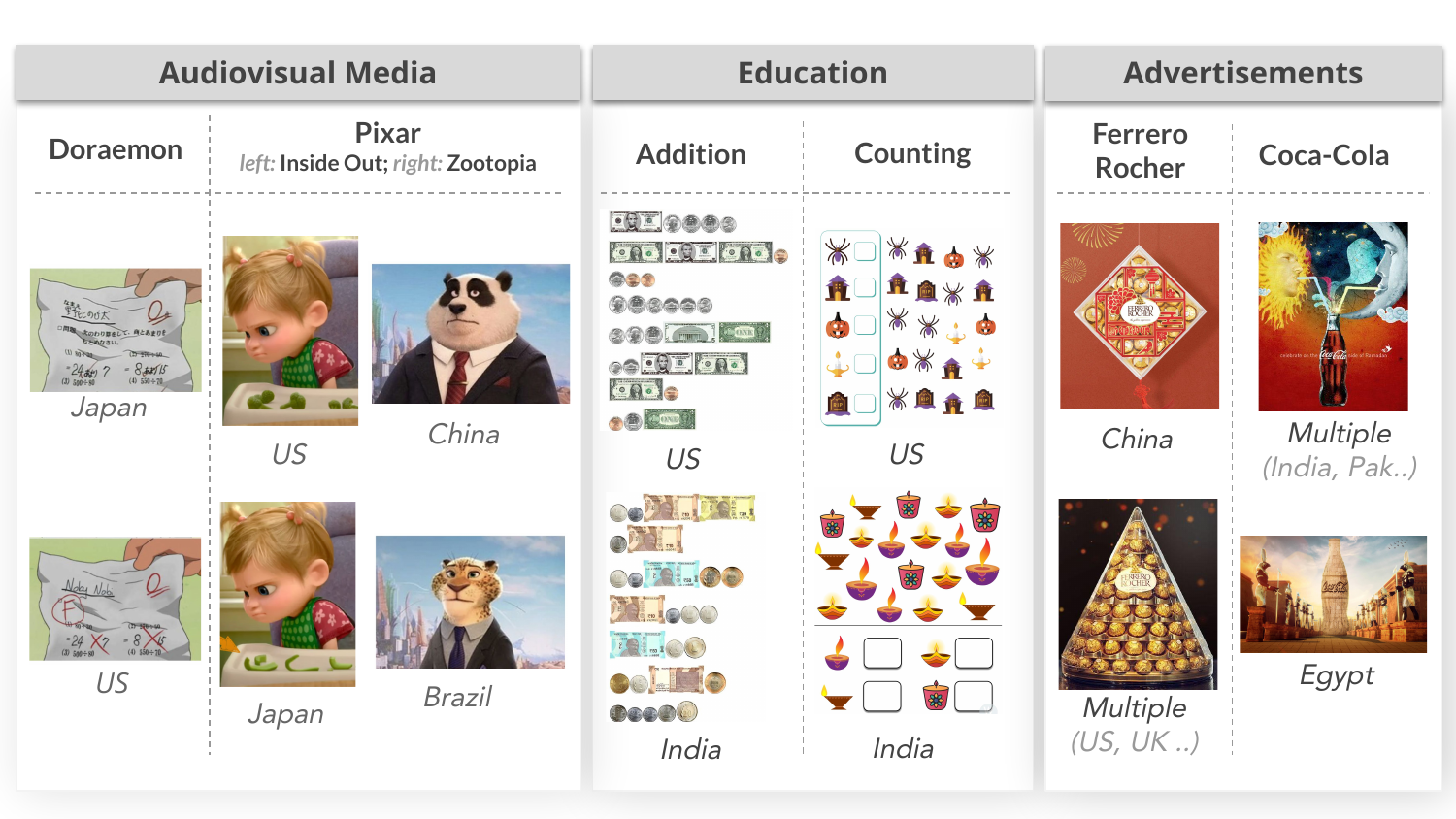}
    \caption{\textbf{Image transcreation} as done in various applications today: \emph{a) Audiovisual (AV) media}: where several changes were made to adapt Doraemon to the US context like adding crosses and Fs in grade sheets, or in Inside Out, where broccoli is replaced with bell peppers in Japan as a vegetable that children don't like; \emph{b) Education}: where the same concepts are taught differently in different countries, using local currencies or celebration-themed worksheets; \emph{c) Advertisements}: where the same product is packaged and marketed differently, like in Ferrero Rocher taking the shape of a lunar festival kite in China, and that of a Christmas tree elsewhere.}
    \label{fig:main}
\end{figure*}

Since the time ancient texts were first translated, philosophers and linguists have highlighted the need for cultural adaptation in the process \cite{jerome_pammachius, ibn_khaldun_muqaddimah, dryden_examen_poeticum, jakobson_linguistic_aspects, nida_correspondence_translating} -- achieving the same ``effect'' on the target audience is essential \cite{nida_correspondence_translating}. Further, with increased consumption and distribution of multimedia content, scholars in translation studies \cite{chaume2018audiovisual,ramiere2010you,sierra2008humor} challenge the notion of simply translating words, highlighting that visuals, music, and other elements contribute equally to meaning. While each modality carries its own information, interaction between modalities creates deeper, emergent meanings. Partial translation disturbs this multimodal interaction and causes cognitive dissonance to the receptor \cite{esser2016media}. Traditionally, translation has been associated with language in speech and text. To broaden its scope to all modalities, and emphasize on the translator's creative role in the process, the term \emph{transcreation} is seeing widespread adoption today.

\emph{Transcreation} is prevalent in several fields and its precise implementation is often tied to the end-application, as shown in Figure \ref{fig:main}. For example, in \emph{audio-visual media} (AV), the goal is to evoke similar emotions across diverse audiences. In line with this goal, the Japanese cartoon \texttt{Doraemon} made many changes like replacing omelet-rice with pancakes, chopsticks with forks and spoons or yen notes with dollar notes, when adapting content for the US.\footnote{\url{http://tinyurl.com/doraemon-us}} Sometimes, the translation is context-dependant, as in the US movie \texttt{Inside Out}, where bell peppers is used as a substitute for broccoli in Japan, as a vegetable that children don't like. In \emph{education}, the goal is to create content that includes objects a child sees in their daily surroundings, known to aid learning \cite{hammond2020ed}. Many worksheets already do this, where the same concepts of addition and counting are taught using different currency notes or celebration-themed worksheets, in different regions. Finally, in \emph{advertisements and marketing}, we see global brands localize advertisements to sell the same product, a strategy proven to boost sales \cite{ho2016translating}. \texttt{Coca-cola} is a famous example, an embodiment of ``Think Global, Act Local'', that tailors its ads to resonate with local cultures and experiences and deeply connect with its audience. 
\vspace{0.5mm}

\textbf{Contribution 1 \emph{(Task)}}: In this paper, we take a first step towards transcreation with machine learning systems, by assessing capabilities of generative models for the task of \textbf{image transcreation} across cultural boundaries. In text-based systems alone, models struggle with translating culture-specific information, like idioms \citep{liu-etal-2023-crossing}. Moreover, to our knowledge, automatically transcreating visual content has previously been unaddressed. 

\textbf{Contribution 2 \emph{(Pipelines)}}: In \S\ref{sec:models}, we introduce three pipelines for this task -- \textbf{a)} \texttt{e2e-instruct} \emph{(instruction-based image-editing)}: that edits images directly following a natural language instruction; \textbf{b)} \texttt{cap-edit} \emph{(caption $\rightarrow$ LLM edit $\rightarrow$ image edit)}: that first captions the image, makes the caption culturally relevant, and edits the original image as per the culturally-modified caption; \textbf{c)} \texttt{cap-retrieve} \emph{(caption $\rightarrow$ LLM edit $\rightarrow$ image retrieval)}: that uses the culturally-modified caption from \texttt{cap-edit} to retrieve a natural image instead. We also experiment with GPT-4o and DALLE-3 to generate new images using culturally-modified captions (\S\ref{app:cap-gen}).


\textbf{Contribution 3 \emph{(Evaluation dataset)}}: Given the unprecedented nature of this task, the evaluation landscape is a blank slate at present. We create an extensive and diverse evaluation dataset consisting of two parts (\emph{concept} and \emph{application}), as detailed in \S\ref{sec:dataset}. \emph{Concept} comprises 600 images across seven geographically diverse countries: Brazil, India, Japan, Nigeria, Portugal, Turkey, and United States. Five culturally salient concepts and related images are collected across a consistent set of universal categories (like food, beverages, celebrations, and so on) from each country. \emph{Application} comprises 100 images curated from real-world applications like educational worksheets and children's literature. 

\begin{figure*}[!htbp]
    \centering
    \includegraphics[width=0.98\textwidth]{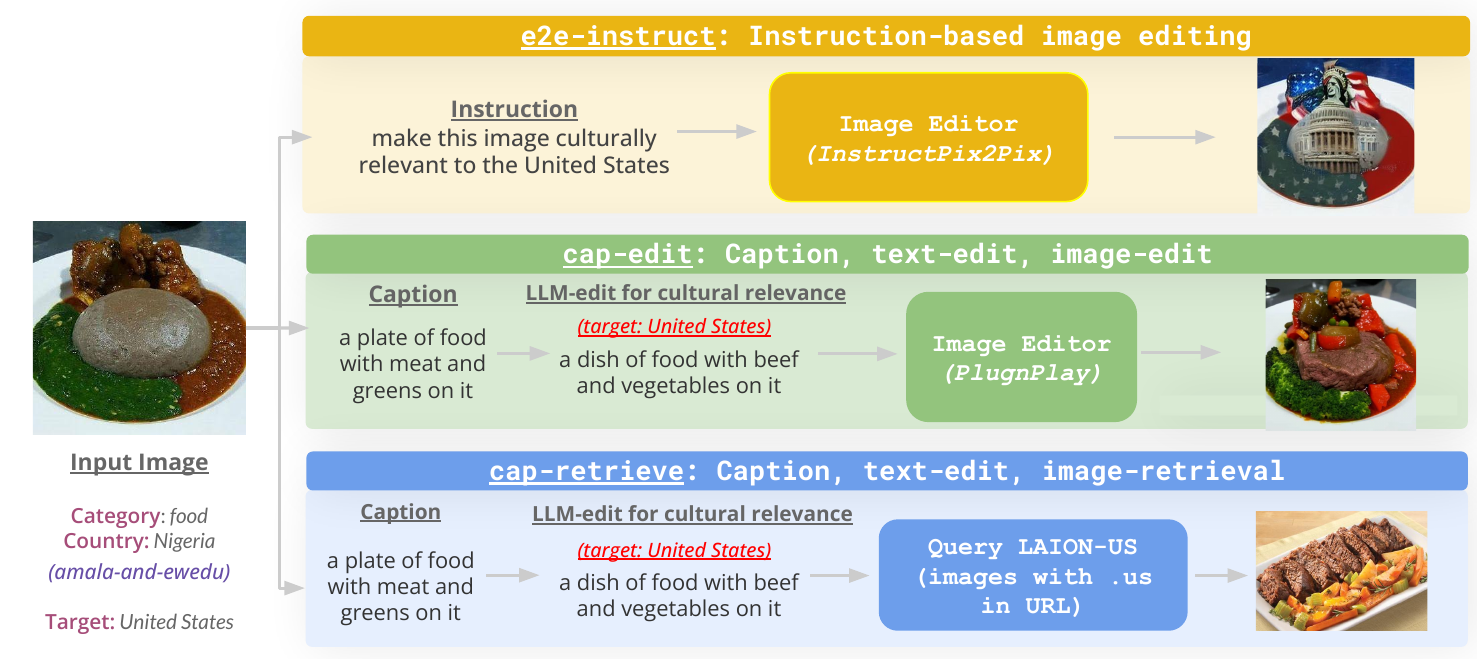}
    \caption{\emph{Pipelines to transcreate images:} \texttt{e2e-instruct} takes as input the original image and a natural language instruction; \texttt{cap-edit} first captions the image, uses a LLM to edit the caption for cultural relevance, and edits the original image using the LLM-edit as instruction; and \texttt{cap-retrieve} uses this LLM-edit to retrieve a natural image from a country-specific image dataset. Given the unprecedented nature of this task, we create pipelines using pre-existing SOTA models, and benchmark them on our newly created test set.}
    \label{fig:models}
\end{figure*}

\textbf{Contribution 4 \emph{(Human evaluation)}}: In \S\ref{sec:results}, we conduct human evaluation of images transcreated for both \emph{concept} and \emph{application}, across all seven countries. We find that as of today, image-editing models fail at this task, but can be improved by leveraging LLMs and retrievers in the loop. Even the best models can only successfully transcreate 5\% images for Nigeria in the simpler \emph{concept} dataset and no image transcreation is successful for some countries in the harder \emph{application} dataset. 

\section{Pipelines for Image Transcreation}
\label{sec:models}
We introduce three pipelines for image transcreation comprising of state-of-the-art generative models. The code to run all pipelines with exact prompts used can be found in Table \ref{tab:prompts}. An overview of each pipeline is in Figure \ref{fig:models}.

\vspace{-1mm}
\subsection{\texttt{e2e-instruct}: Instruction-based editing}
First, we use out-of-the-box instruction-based image editing models to translate the image in one pass. Specifically, we use InstructPix2Pix \cite{brooks2023instructpix2pix}, a model that allows users to define edits using natural language, as opposed to other models requiring text labels, captions, segmentation masks, example output images and so on.\footnote{\url{https://www.timothybrooks.com/instruct-pix2pix}} 

We feed in the original image and instruct the model to \emph{make the image culturally relevant to \texttt{COUNTRY}}, following a similar prompt format as that used to train the model. This pipeline is simple and flexible, but relies heavily on the image models' ability to perform culturally relevant edits, which it is currently incapable of doing, as discussed in \S\ref{sec:results}.



\subsection{\texttt{cap-edit}: Caption, text-edit, image-edit}

Our second approach is a modular pipeline that offloads some of the requirement of cultural understanding from image editing models to large language models (LLMs). LLMs have been trained on trillions of tokens of text \cite{touvron2023llama,achiam2023gpt}, and exhibit at least a certain degree of cultural awareness \cite{arora2022probing}. Concretely, we adopt a method that first performs image captioning, edits the caption for cultural relevance using an LLM, and then edits the image using an instruction-based image editing model.
In experiments, we use InstructBLIP-FlanT5-XXL\footnote{\url{https://huggingface.co/Salesforce/instructblip-flan-t5-xxl}} \citep{li2023blip} as the image captioner, GPT-3.5\footnote{\url{https://platform.openai.com/docs/models/gpt-3-5}} for caption transformation, and PlugnPlay as the image editing model \cite{tumanyan2023plug}.



\begin{figure*}[!ht]
    \centering
    \includegraphics[width=0.98\textwidth]{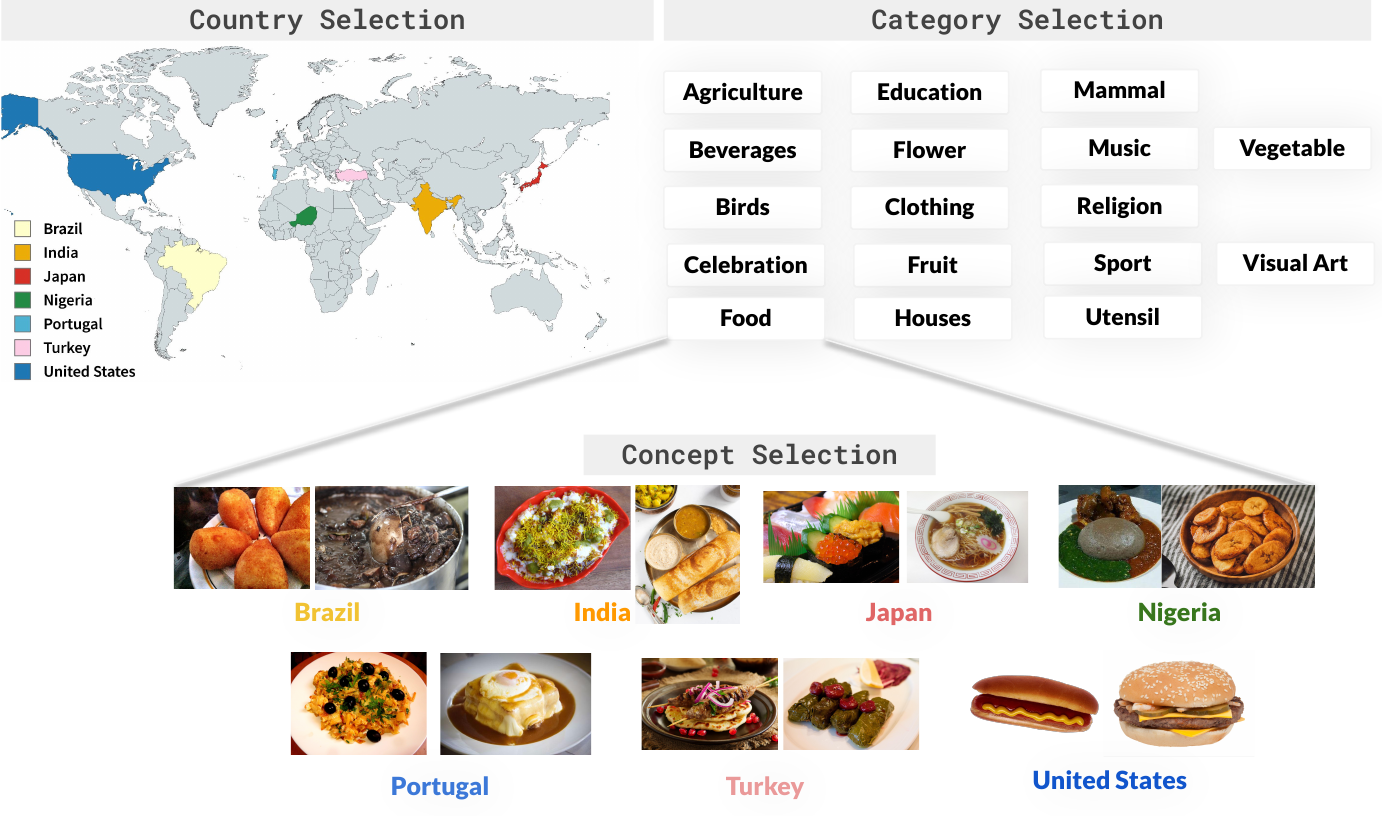}
    \caption{\emph{Concept} dataset: We select seven geographically diverse countries and universal categories that are cross-culturally comprehensive. Annotators native to selected countries give us 5 concepts and associated images that are culturally salient for the speaking population of their country.}
    \label{fig:eval-part1}
\end{figure*}

\subsection{\texttt{cap-retrieve}: Caption, edit, retrieve}
In \texttt{cap-edit}, the final output is sometimes not reflective of how the concept naturally appears in the target country, due to image-editing models being trained to strictly preserve spatial layout (\S\ref{app:cap-edit}). Hence, here we rely on retrieval from a country-specific image database instead. Concretely, we first caption the image and edit the caption for cultural relevance, similar to \texttt{cap-edit}. Next, we use the LLM-edited caption to query country-specific subsets of LAION \cite{schuhmann2022laion}. These subsets are created by parsing image URLs and categorizing them based on the country-code top-level domain they contain. For example, URLs featuring ``.in'' are assigned to the India subset, those with ``.jp'' are grouped into the Japan subset, etc.

\section{Evaluation Dataset}
\label{sec:dataset}
We design a two-part dataset where the first (\emph{concept}) is meant to serve as a research prototype, while the second (\emph{application}) is grounded in real-world applications like those in Figure \ref{fig:main}.

\subsection{\emph{Concept} dataset}
We collect images for a set of universal categories, across seven countries (Figure \ref{fig:eval-part1}). We follow the annotation protocol of MaRVL \cite{liu2021visually} for which people local to a region drive the entire annotation process, ensuring the collected data accurately captures their lived experiences. Concretely, our collection process is as follows:

\noindent \textbf{Country Selection:} We select seven geographically diverse countries: Brazil, India, Japan, Nigeria, Portugal, Turkey, and United States. But do geographic borders dictate cultural ones? Cultures constantly change and are hybrid at any point in time \cite{hall2015cultural}. However, audiovisual adaptation is most often equated with national boundaries \cite{moran2009global, keinonen2016cultural}, given the significant influence of history, policy, and state regulations on media consumption within countries \cite{steemers2012evaluating}. Further, from a practical perspective, ML systems need data, whose source can be geographically tagged and segregated. While the ultimate goal is to adapt to individual experiences that shape cultural contexts, focusing on the national level serves as a practical starting point.

\noindent \textbf{Category Selection:} Ideally, datasets for different cultures should reflect most salient concepts as they naturally occur in that culture, while retaining some thematic coherence for comparability \citet{liu2021visually}. Hence, we opt for a list of universal concepts that are cross-culturally comprehensive, as laid out in the Intercontinental Dictionary Series \cite{key2015ids}.

\noindent \textbf{Concept Selection}: We hire five people who are intimately familiar with the culture of each of the countries above, and ask them to list five culturally salient concepts, such that they are \textbf{a)} commonly seen or representative in the speaking population of the language; and \textbf{b)} ideally, are physical and concrete (details in \S\ref{app:annot}). Aggregating all responses, we retain top-5 most frequent concepts in each category, for each country.

\noindent \textbf{Post-Filtering}: The selected concepts and images are additionally verified by 3 native speakers, and those without a majority voting ($<$ 2) are filtered out. We obtain 85 images per country, which become roughly 580 images overall, post-filtering.



\subsection{\emph{Application} dataset} 
\label{sec:part2-eval}
The second part of the dataset is curated from real-world applications (\emph{education} and \emph{literature}), a choice guided by availability of data resources.

\noindent \textbf{Education: } Research suggests that incorporating 
objects in a child’s surrounding and grounding content in their culture aids learning \cite{national2015transforming}. Looking at math worksheets for grades 1-3, we find this to be true. We source worksheets from K5 Learning,\footnote{\url{https://www.k5learning.com/free-worksheets-for-kids} We obtain permission to use and distribute the worksheets for non-commercial research purposes from the publisher.} a US-based learning platform. The transcreation process is tied to the task here, and may not be as straightforward as replacing currency notes in Figure \ref{fig:main}. For example, in the left below, the model must find differently-colored elements while retaining the count of each colored object during transcreation, or on the right, where its necessary to find objects that can be measured using the chosen replacement for a matchstick.

\begin{figure}[htbp]
  \centering
  \begin{subfigure}[b]{0.22\textwidth}
    \includegraphics[width=0.98\textwidth]{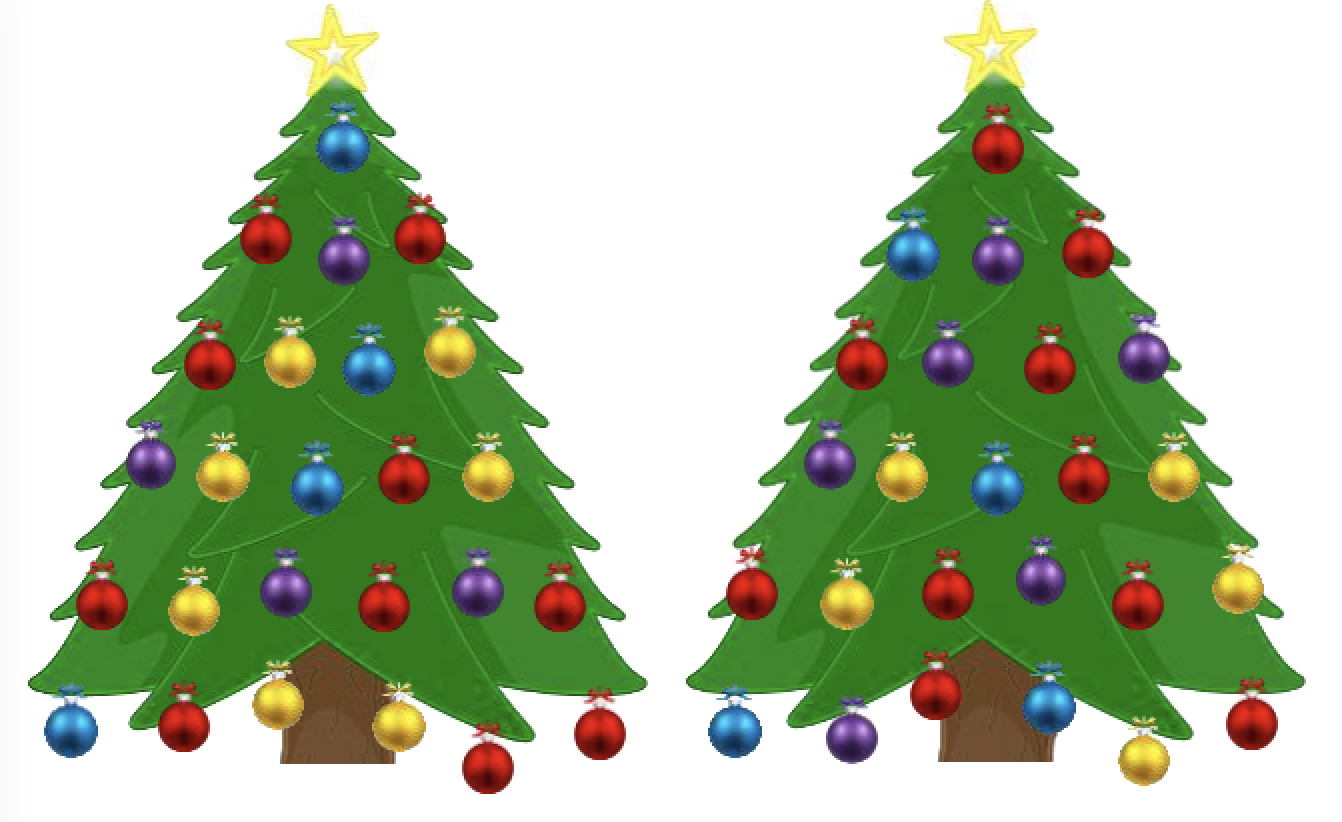}
    \caption{\emph{Task (counting):} Count the number of blue, violet, red and yellow christmas balls}
    \label{fig:christmas}
  \end{subfigure}
  \hfill
  \begin{subfigure}[b]{0.24\textwidth}
    \includegraphics[width=0.98\textwidth]{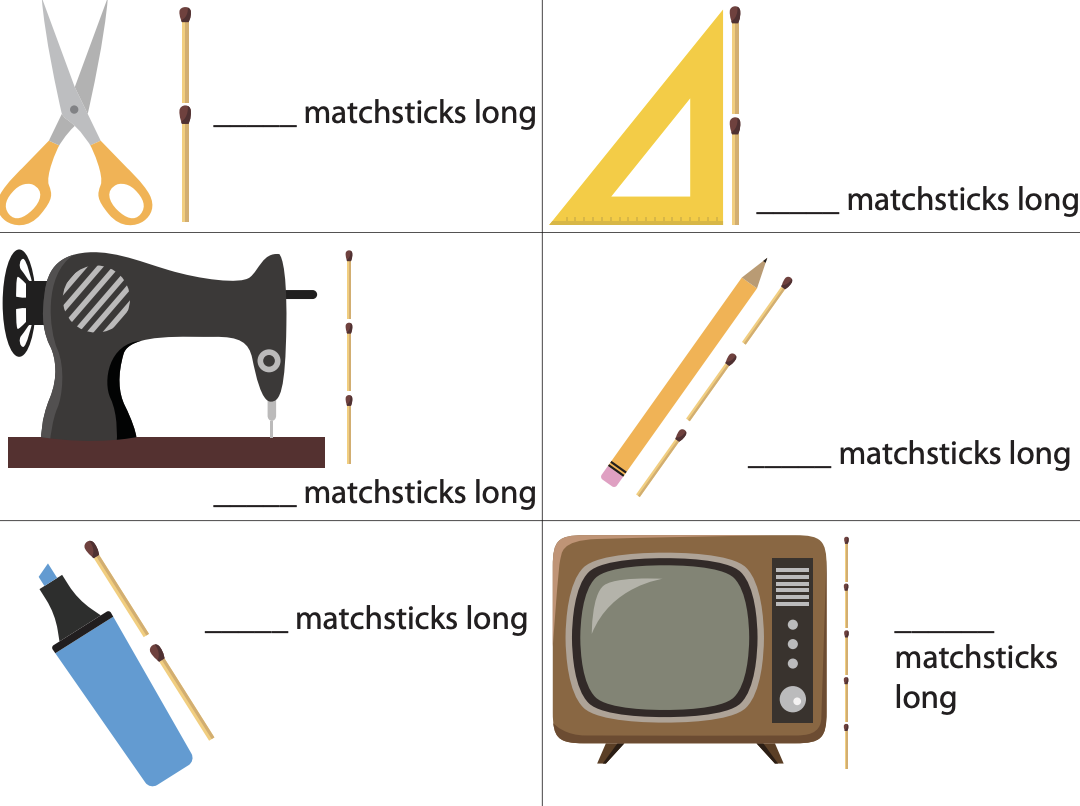}
    \caption{{\emph{Task (measurement):} Use matchsticks to measure objects}}
    \label{fig:matchstick}
  \end{subfigure}
  \label{fig:eval-part2}
\end{figure}

\begin{figure}[h]
    \centering
    \includegraphics[width=0.4\textwidth]{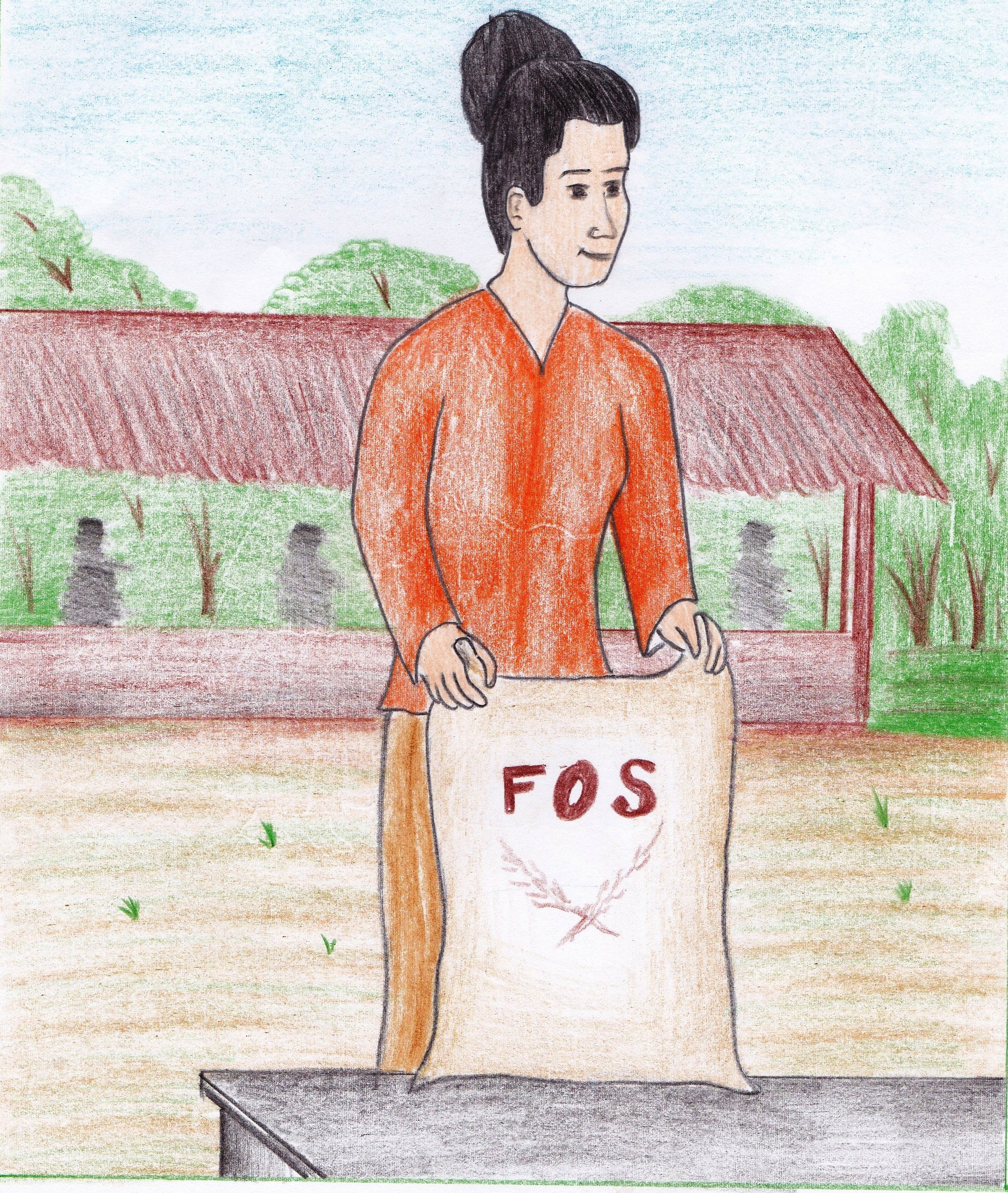}
    \caption{\emph{Story text:} My mom bought rice.}
    \label{fig:story}
    \vspace{-4mm}
\end{figure}

\noindent\textbf{Literature:} We curate images from Bloom Library,\footnote{\url{https://bloomlibrary.org/}} a digital library of stories for children released for research purposes by \citet{leong2022bloom}.\footnote{\url{https://huggingface.co/datasets/sil-ai/bloom-vist}} Dealing with a sequence of images is out-of-scope of our current work, hence we collect the first image in each story along with its text that is later used to guide the transcreation. We manually select roughly 60 images out of 400 from the \emph{eng} subset, making sure the selected images are of high quality and de-duplicated (\reffig{fig:story}).

\subsection{Why the two-part dataset?}
Even though our eventual goal is to transcreate images for real-world applications, real-world scenes are complex, comprising of multiple interacting objects, and have application-specific constraints, making the task harder. For example, in Figure \ref{fig:matchstick}, one is constrained to find objects of a specific length that can be measured using a matchstick. 

With \emph{concept}, we build a prototype which has the following features: \textbf{a)} \emph{diverse}: images are collected across 7 geographically spread-out countries; \textbf{b)} \emph{single concept or object per image}: making it easier to analyse model errors when one image represents a concept in isolation; \textbf{c)} \emph{loose constraints on output}: the goal is simply to increase cultural relevance while staying within bounds of the universal category.


Below, we discuss how all models face difficulties even with \emph{concept}, further strengthening the need for it in evaluation.

\begin{table*}
  \centering
  \small
  \begin{tabular}{|p{0.42cm}|p{5.7cm}|p{2cm}|p{2cm}|p{3.5cm}|}
    \hline
    \textbf{ID} & \textbf{Question} & \textbf{Property} & \textbf{Applications} & \textbf{Performance} \\
    \hline
    \multicolumn{5}{|c|}{\textbf{Concept Dataset}} \\
    \hline
    C0 & Is there any visual change in the generated image compared to the original image? & \texttt{visual- change} & None (\emph{helps filter non-edits}) & \hllightblue{\texttt{e2e-instruct}} \hlmedblue{\texttt{cap-edit}} \hldarkblue{\texttt{cap-retrieve}}  \\
    \hline
    C1 & Is the generated image from the same semantic category as the original image? & \texttt{semantic- equivalence} & AV (Zootopia); Education & \hllightblue{\texttt{e2e-instruct}} \hldarkblue{\texttt{cap-edit}} \hlmedblue{\texttt{cap-retrieve}} \\
    \hline
    C2 & Does the generated image maintain spatial layout of the original image? & \texttt{spatial- layout} & AV (Doraemon, Inside Out) & \hlmedblue{\texttt{e2e-instruct}} \hldarkblue{\texttt{cap-edit}} \hllightblue{\texttt{cap-retrieve}}\\
    \hline
    C3 & Does the image seem like it came from your country/ is representative of your culture? & \texttt{culture- concept} & AV, Education, Ads & \hllightblue{\texttt{e2e-instruct}} \hlmedblue{\texttt{cap-edit}} \hldarkblue{\texttt{cap-retrieve}} \\
    \hline
    C4 & Does the generated image reflect naturally occurring scenes/objects? & \texttt{naturalness} & Ads (Ferrero Rocher) & \hllightblue{\texttt{e2e-instruct}} \hlmedblue{\texttt{cap-edit}} \hldarkblue{\texttt{cap-retrieve}} \\
    \hline
    C5 & Is this image offensive to you, or is likely offensive to someone from your culture? & \texttt{offensiveness} & All & \hlmedblue{\texttt{e2e-instruct}} \hllightblue{\texttt{cap-edit}} \hldarkblue{\texttt{cap-retrieve}}  \\
    \hline
    - & For edited images, is the change meaningful (C1) and culturally relevant (C3)? & \texttt{meaningful- edit} & All & \hllightblue{\texttt{e2e-instruct}}  \hlmedblue{\texttt{cap-edit}} \hldarkblue{\texttt{cap-retrieve}} \\
    \hline
    \multicolumn{5}{|c|}{\textbf{Application Dataset}} \\
    \hline
    E/S0 & Is there any visual change in the generated image compared to the original image? & \texttt{visual- change} & None (\emph{helps filter non-edits}) & \hllightblue{\texttt{e2e-instruct}} \hlmedblue{\texttt{cap-edit}}
    \hldarkblue{\texttt{cap-retrieve}}  \\
    \hline
    E1 & Can the generated image be used to teach the concept of the worksheet? & \texttt{education- task} & Education & \hllightblue{\texttt{e2e-instruct}} \hldarkblue{\texttt{cap-edit}} \hlmedblue{\texttt{cap-retrieve}}  \\
    \hline
    S1 & Would the generated image match the text of the story in a children’s storybook? & \texttt{story-text} & AV, Literature & \hllightblue{\texttt{e2e-instruct}} \hldarkblue{\texttt{cap-edit}} \hlmedblue{\texttt{cap-retrieve}} \\
    \hline
    E/S2 & Does the image seem like it came from your country/is representative of your culture? & \texttt{culture- application} & All & \hllightblue{\texttt{e2e-instruct}} \hlmedblue{\texttt{cap-edit}} \hldarkblue{\texttt{cap-retrieve}} \\
    \hline
    - & For edited images, is the change meaningful (E/S1) and culturally relevant (E/S2)? & \texttt{meaningful- edit} & All & \hllightblue{\texttt{e2e-instruct}} \hldarkblue{\texttt{cap-edit}} \hlmedblue{\texttt{cap-retrieve}} \\ 
    \hline
  \end{tabular}
  \caption{Questions asked for evaluation, the applications a model with this property would benefit (examples from Figure \ref{fig:main}), and the pipeline ranking for the property tested (\hldarkblue{first} \hlmedblue{second} \hllightblue{third}).}
  \label{tab:qs_part1}
\end{table*}

\section{Human Evaluation and Quantitative Metrics}
\label{sec:results}
Evaluation of image-editing models typically relies on quantitative metrics and qualitative analysis of a few select samples.\footnote{Some skip a quantitative evaluation altogether as in \citet{hertz2022prompt}.} While image-editing focuses on image quality and how closely the edit follows the instruction, image-transcreation comes with additional requirements such as cultural relevance, meaning preservation, and so on. Hence, we design an extensive questionnaire and conduct human evaluation to assess the quality of \emph{all} generated images, across both parts of the dataset (Table \ref{tab:qs_part1}). Evaluators are shown the source image and the three pipeline outputs in a single instance, (Figure \ref{fig:screenshot}). This ensures that scores capture relative differences across pipelines. Further, the order of pipeline outputs is randomized so as to not bias the ratings. 



\subsection{\emph{Questions and Findings}: Concept}
\label{sec:proto_qs}

\textbf{End Goal}: To transcreate the image such that the final image: \textbf{a)} belongs to the same universal category as the original (like food, animals etc.), and \textbf{b)} has higher cultural relevance than the original image, for a given target country.

\begin{figure*}
    \centering
    \includegraphics[width=\linewidth]{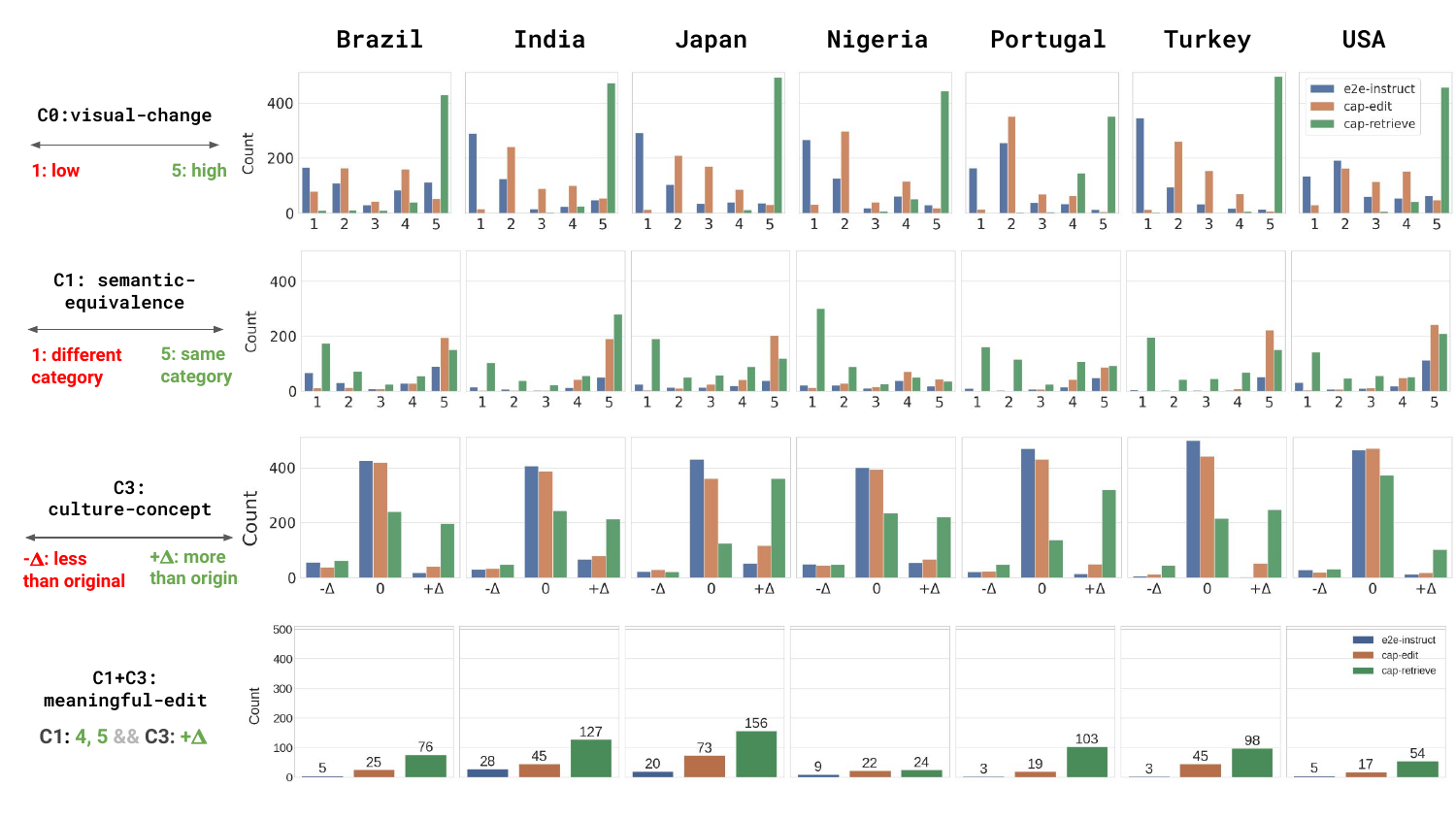}
    \caption{\emph{Human ratings for the concept dataset}: Our primary goal is to test whether the \emph{edited image belongs to the same universal category} as the original image (\textbf{C1}) and whether it \emph{increases cultural relevance} (\textbf{C3}). We plot the count of images that can do both above (\textbf{C1+C3}), and observe that the best pipeline's performance ranges between 5\% (Nigeria) to 30\% (Japan).}
    \label{fig:concept-result}
\end{figure*}

\noindent However, note that we ask many more questions on layout preservation, offensiveness etc, since different applications may have different constraints on the output, as shown in Table \ref{tab:qs_part1}. A summary of responses are below, while detailed analyses of responses can be found in \S\ref{app:human_eval}: 

\textbf{C0: Is there any visual change in the generated image, when compared with the source image?} 
\texttt{cap-retrieve} maximally edits images, with roughly 90\%  scoring \texttt{5} (Figure \ref{fig:concept-result}); \texttt{e2e-instruct} makes no edit sometimes, with 40-60\% images scoring \texttt{1}; and \texttt{cap-edit} lies mid-way.

\textbf{C1: If an edit is made, is it meaningful?} For images with \textbf{C0 $>2$}, (indicating some visual changes), we observe that \texttt{cap-edit}'s changes maximally retain the universal category, for ex., a food item from country A is changed to another food item from country B; whereas  \texttt{e2e-instruct} often makes meaningless edits like pasting flag colors of the target country on the image (\S\ref{app:e2e-instruct}). \texttt{cap-retrieve} is highly variable; for some countries (India, US), it is better than \texttt{cap-edit} and for some (Nigeria), it is very noisy.


\textbf{C3: Are the edited images more culturally relevant than the original image?} Here, we compare the change in the final image's cultural relevance score with the original image (Figure \ref{fig:concept-result}). \texttt{cap-retrieve} has the highest \% of images with a positive change, followed by \texttt{cap-edit} after a relatively large gap, while \texttt{e2e-instruct} performs worst. This shows that offloading the cultural translation to LLMs generally helps, and natural images are highly preferred over edited images when assessing for culture.

\textbf{C1+C3: What proportion of images are successfully transcreated?} We define \textbf{C0 $>2 \And$ C1 $>2 \And$ C3$_{edited}>$ C3$_{original}$} as the criteria for a successful transcreation. Best pipelines can only transcreate 5\% images for some countries (Nigeria); while the accuracy is 30\% for some others (Japan), indicating that this task is far from solved.


\subsection{\emph{Questions and Findings}: Application}
\label{sec:appl_qs}
\textbf{End Goal (\emph{Education})}: To transcreate such that the final image: \textbf{a)} can be used to teach the same concept as the original image (like counting); \textbf{b)} has higher cultural relevance than the original image, for a given target country.

\noindent \textbf{End Goal (\emph{Stories})}: To transcreate such that the final image: \textbf{a)} matches the text of the story; \textbf{b)} has higher cultural relevance than the original image, for a given target country.

\begin{figure*}[ht!]
    \centering
    \begin{subfigure}[b]{0.22\textwidth}
        \includegraphics[width=\textwidth]{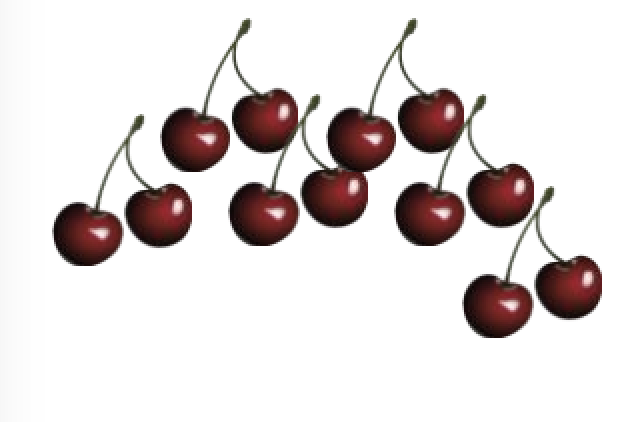}
        \caption{original}
    \end{subfigure}
    \hfill 
    \begin{subfigure}[b]{0.22\textwidth}
        \includegraphics[width=\textwidth]{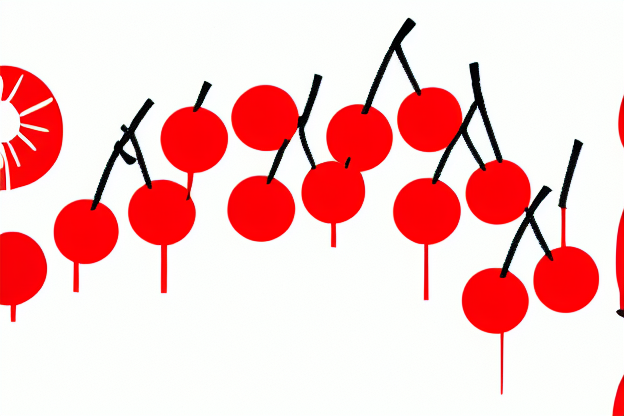}
        \caption{\texttt{e2e-instruct}}
    \end{subfigure}
    \hfill 
    \begin{subfigure}[b]{0.22\textwidth}
        \includegraphics[width=\textwidth]{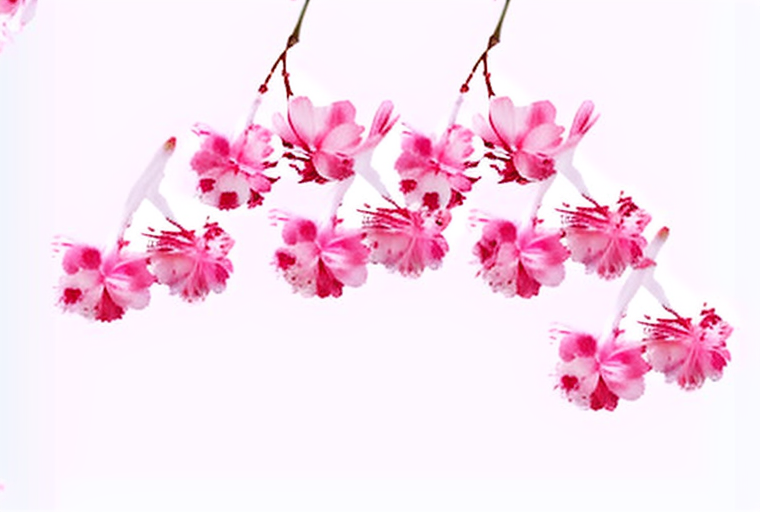}
        \caption{\texttt{cap-edit}}
    \end{subfigure}
    \hfill 
    \begin{subfigure}[b]{0.22\textwidth}
        \includegraphics[width=\textwidth]{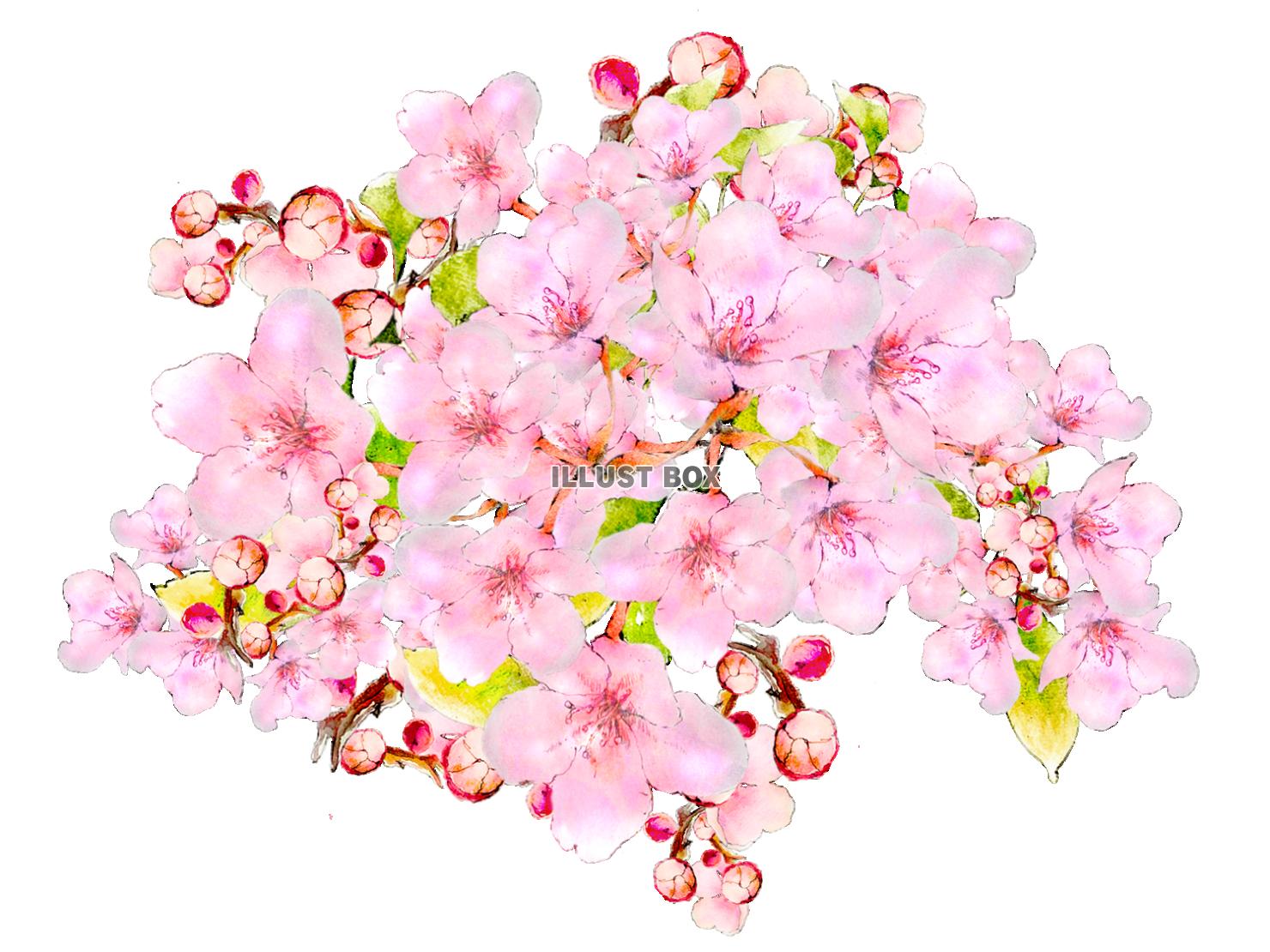}
        \caption{\texttt{cap-retrieve}}
    \end{subfigure}
    \caption{\emph{Application:} Education; \emph{Target:} Japan --- \emph{Task}: count the number of cherries. \texttt{cap-edit} is a successful transcreation despite the semantic drift from a fruit to a flower, because the final image can be used to teach counting to children.}
    \label{fig:app_ed}
\end{figure*}

\noindent \textbf{Observations}: Overall, responses to individual questions are similar to as observed for the \texttt{concept} dataset. The task here is much harder than simply transcreating within a universal category like in \texttt{concept} because of which no image is successfully transcreated by any pipeline for some countries (Portugal). In Figure \ref{fig:app_ed} we see a sample output where \texttt{e2e-instruct} makes the cherries a red that resembles the Japan flag, and \texttt{cap-edit} is a successful transcreation because even though there is a semantic drift from cherries to flowers, the worksheet can be used to teach counting.  Detailed results are in \S\ref{app:appl}.

\subsection{Quantitative Metrics}
For image-editing, these typically capture how closely the edited image matches -- \textbf{(i)} the original image; and \textbf{(ii)}  the edit instruction. Following suit, we calculate two metrics:

\noindent \textbf{a)} \emph{image-similarity}: we embed the original image and each of the generated images using DiNO-ViT \cite{caron2021emerging} and measure cosine similarity

\noindent \textbf{b)} \emph{country-relevance}: we embed the text -- \texttt{This image is culturally relevant to \{COUNTRY\}}, and the edited images using CLIP \cite{radford2021learning} and calculate their cosine similarity. 

We present results for both metrics in Figures \ref{fig:target_source_sim} and \ref{fig:image_sim}. A discussion on correlation of these metrics with human evaluation is in \S\ref{app:quant_eval}.

We find that overall for \emph{image-similarity}, \texttt{e2e-instruct} scores highest, closely followed by \texttt{cap-edit}, while \texttt{cap-retrieve} lags behind, consistent with human ratings. However, note that our goal here is to have the right trade-off between image-similarity and the naturalness of the edited image (which cannot be captured by this metric). Figure \ref{fig:chai} shows an example of the final image having a high similarity with the original, but nonetheless looks unnatural.

For the \emph{country-relevance score}, we observe that it has a high recall but low precision. These scores are positively correlated with human ratings for \textbf{C3:} \texttt{cultural-relevance}, but this metric also scores images containing stereotypical artifacts (such as the ones discussed in \S\ref{app:e2e-instruct}) high on cultural relevance.

Our findings above indicate that quantitative metrics cannot sufficiently capture the quality of transcreation of an image, and developing a BLEU-equivalent, but for images, would be necessary to make measurable progress on this task.


\section{Related Work}

\noindent \textbf{Cultural diversity in image generation}: Several recent works investigate cultural awareness of text-to-image (T2I) systems typically highlighting biases towards certain cultures. \citet{hutchinson2022underspecification} highlight how under-specified prompts show gender and western cultural biases, \citet{jha2024visage} analyse regional stereotypical markers in generated images, \citet{naik2023social} discuss occupational biases of neutral prompts and personality trait associations with limited groups of people, \citet{cho2023dall} reveal skin-tone biases and \citet{bird2023typology} discuss associated risks of these biases for society. Some other works focus on ways to probe for and evaluate cultural relevance of generated images. \citet{ventura2023navigating} derive prompt templates to unlock the cultural knowledge in T2I systems, and \citet{hall2023dig} evaluate the realism and diversity of T2I systems when prompted to generate objects from across the world. While all of these works are targeted towards assessing and mitigating cultural biases in pre-trained models, our work is targeted towards an \emph{application} (i.e. transcreating visual content) that would benefit by such efforts that improve the cultural understanding and diversity of image generation models. \\

\noindent \textbf{Image-editing models} have evolved over the years from being capable of single editing tasks like style transfer \cite{gatys2015neural, gatys2016image} to handling multiple such tasks in one model \cite{isola2017image,choi2018stargan,huang2018multimodal,ojha2021few}. Today, their capabilities range from performing targeted editing that preserves spatial layout, local in-painting, to edits that can follow natural language instructions \cite{brooks2023instructpix2pix}. We choose InstructPix2Pix \cite{brooks2023instructpix2pix} to experiment with, given its flexibility to prompt with natural language instructions, as opposed to other models requiring text labels, captions, segmentation masks, example output images and so on. It has also consistently been one of the most downloaded image-editing models on HuggingFace.\footnote{\url{https://huggingface.co/models?pipeline_tag=image-to-image&sort=downloads}} As discussed in Section \ref{sec:results} however, these models are only capable of making color, shape and style changes, and lack a deeper understanding of natural language. No image-editing works have tackled the semantically complex task of cultural transcreation. We hope that our work paves the way to building image-editing models that truly understand natural language, which can benefit multiple applications, including ours.


\section{Conclusion}
In this paper, we introduce a new task of \textbf{image transcreation} with machine learning systems, where we culturally adapt visual content to suit a target audience. Translation has traditionally been limited to language, but with increased consumption of multimedia content, translating \emph{all} modes in a coherent way is essential. We build three pipelines comprising state-of-the-art generation models, and show that end-to-end image editing models are incapable of understanding cultural contexts, but using LLMs and retrievers in the loop helps boost performance. We create a challenging two-part evaluation dataset: (i) \emph{concept} which is simple, cross-culturally coherent, and diverse; and (ii) \emph{application} which is curated from education and stories. We conduct an extensive human evaluation and show that even the best models can only translate 5\% images for select countries (like Nigeria) in the easier \emph{concept} dataset and no image transcreation is successful for some countries (like Portugal) in the harder \emph{application} dataset. Our code and data is released to facilitate future work in this new, exciting line of research.
\section{Limitations}
\noindent\textbf{Categorizing culture based on country}: 
In \S\ref{sec:dataset}, we acknowledge that cultures do not follow geographic boundaries. It varies at an individual level and is shaped by one's own life experiences. However, the content of several multimedia resources is often influenced by state regulations and policies decided at the national level. Further, a nation has long history which ties people together and influences their languages, customs and way of life. Finally, from a practical standpoint, data for machine learning systems can be segregated based on physical boundaries by geo-tagging it. All these factors convinced us that approaching this problem from a nation-level would be a good starting point. Eventually, we'd like to build something that can learn from individual user interaction, and adapt to varied and ever-evolving cultures. \\

\noindent\textbf{Limited coverage of languages and countries under study: } In this work, we consider seven geographically diverse countries given time and budget constraints involved in data collection and human evaluation. Our choices were also motivated by availability of annotators on the crowd-sourcing platform we use, Upwork. Further, in \texttt{cap-edit} and \texttt{cap-retrieve}, we only explore captioning in English. This is because most image-editing models and retrieval-based models only work with English instructions. However, captioning and querying in languages associated with cultures the images are taken from is certainly an interesting direction for future research. \\

\noindent\textbf{A one-to-one mapping may never exist:} One may argue that a perfect substitute or equivalent of an object in another culture may never exist. While this is certainly true, we'd like to highlight that our focus here is on context-specific substitutions that convey the intended meaning within a localized setting. For example, in Figure \ref{fig:main}, we observe that \emph{Inside Out} substitutes broccoli with bell peppers in Japan to convey the concept of a disliked vegetable. However, in the absolute sense, bell peppers is not a substitute for broccoli when we consider other properties like taste, texture, etc. Importantly, the goal of transcreation is to, at the least, \emph{increase} the relatability of the adapted message when compared with the original message. This is also the reason why we compare between the original and edited image's cultural relevance score in the human evaluation in \S\ref{sec:results}, rather than simply looking at absolute cultural relevance values of edited images. \\

\section{Ethical Considerations}

\noindent\textbf{What is the trade-off between relatability and stereotyping?} Often times, models may be prone to stereotyping and only producing a small range of outputs when instructed to increase cultural relevance. We observe this a lot with InstructPix2Pix, where it randomly starts inserting sakura blossoms and Mt. Fuji peaks, out of context, to increase cultural relevance for Japan. Hence, it is essential that we build models capable pf producing a diverse range of outputs while not propagating stereotypes. Importantly, one must note that the problem itself \emph{does not} suggest promoting stereotypes but rather an output that the audience can relate to better. We must move towards developing solutions that enable one to hit any of the multiple possible right answers in their context. \\

\noindent\textbf{We may want to preserve the original cultural elements at times: }
We are also aware that many a times, the goal may be to expose the audience to diverse cultural experiences and not to localize. While we acknowledge that this is extremely important for sharing knowledge and experiences, our work is not applicable in such scenarios. It may also be that we may want to preserve certain elements, while adapt others. In the Japanese anime \emph{Doraemon} for example, creators make some edits to adapt to the US, but preserve most of the original content which is set in the Japanese context. In future work, we'd ideally want to build a system that allows us to visit different points in the relatability/preservation spectrum, that provides for finer-grained object-level control in translation. \\ 

\noindent\textbf{Using pre-existing material created for educational and literary purposes: } Our application-oriented evaluation dataset is curated from content originally created to teach math concepts (education) or for children's literature. The StoryWeaver images are CC-BY-4.0 licensed, and we have been in communication with the team for simpler curation and release of data for the future. There were no licenses associated with educational worksheets. Hence, we obtain written consent to use and distribute their worksheet for non-commercial academic research purposes only. The written consent is obtained for the following task description and purpose: 

\emph{Description of Task}: We are assessing the capabilities of generative AI technology to edit images and make them more relevant to a particular culture. There are many concepts that are culture-specific, which people who have not been immersed in the culture may not understand or be aware of. An important end-application where something like this would be useful is education. For example, if one wants to adapt this math worksheet for children in Japan\footnote{\url{https://www.k5learning.com/worksheets/math/data-graphing/grade-1-same-different-c.pdf}}, they might want to replace Christmas trees with Kadomatsu (bamboo decorations used on new years). We found several such worksheets which could benefit from such local adaptation.

\emph{Purpose of Use}: This is a non-commercial research project. We wish to use some of these images (complete list below), to evaluate our pipelines on cultural adaptation. We also request for permission to distribute to other researchers for non-commercial research purposes only. Please note that we are not training any model on this data and it is being used for testing purposes only. Additionally, if you find our research to be beneficial to your workflow, we would be happy to discuss long-term engagements and collaboration as well.

\section{Acknowledgements}
We would like to thank Shachi Dave, Sagar Gubbi, Daniel Fried, Fernando Diaz, Utsav Prabhu, Jing Yu Koh, Michael Saxon, Frederic Gmeiner, Jeremiah Milbauer, Vivek Iyer, Faria Huq, and members of Neulab for helpful feedback on this work! This work was supported in part by grants from Google and the Defense Science and Technology Agency Singapore.

\bibliography{custom}

\appendix
\label{sec:appendix}
\section{Example Outputs}
Here, we include sample outputs from the pipelines for select images. All pipelines have their own set of limitations, indicating that we have a long way to go before we can solve this task. Patterns observed for each pipeline can be found below: 

\subsection{\texttt{e2e-instruct}: Instruction-based editing}
\label{app:e2e-instruct} 
\begin{figure*}[ht!]
    \centering
    \begin{subfigure}[b]{0.22\textwidth}
        \includegraphics[width=\textwidth]{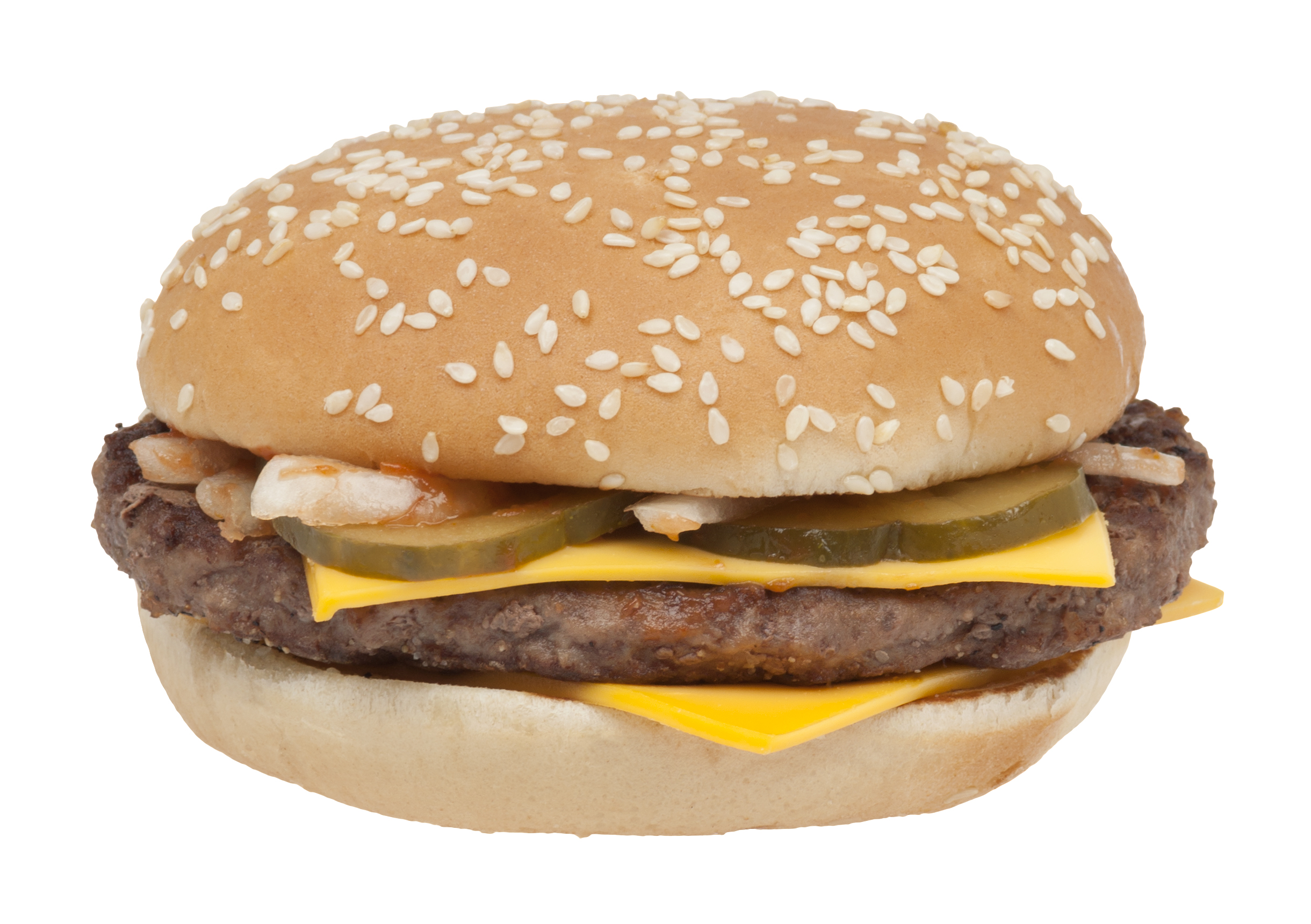}
        \caption{original}
    \end{subfigure}
    \hfill 
    \begin{subfigure}[b]{0.22\textwidth}
        \includegraphics[width=\textwidth]{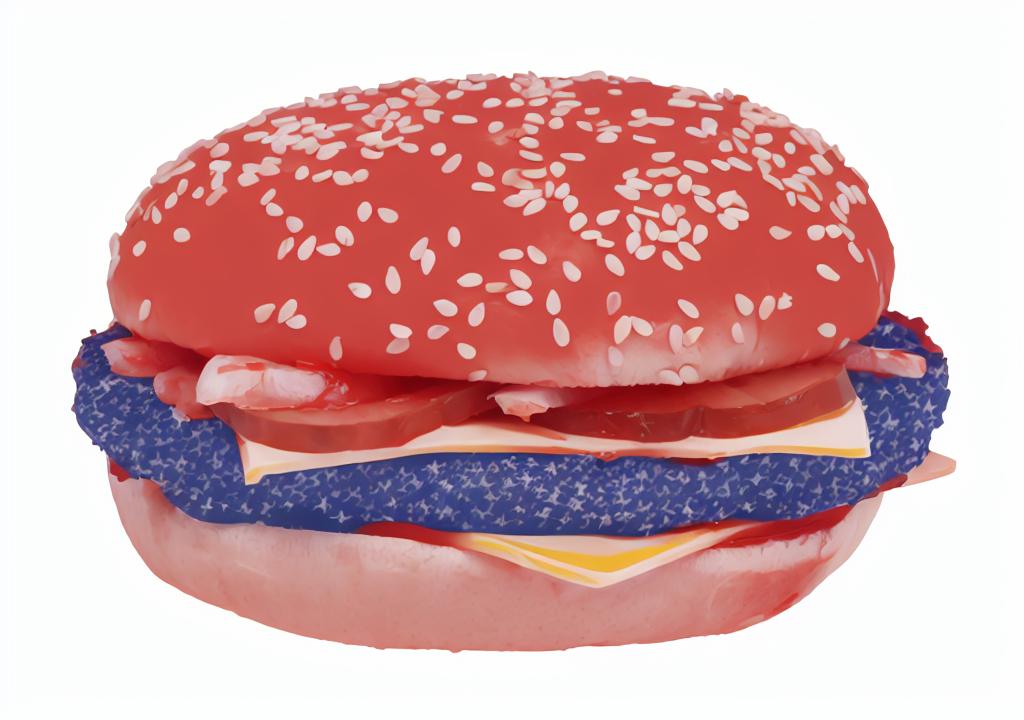}
        \caption{\texttt{e2e-instruct}}
    \end{subfigure}
    \hfill 
    \begin{subfigure}[b]{0.22\textwidth}
        \includegraphics[width=\textwidth]{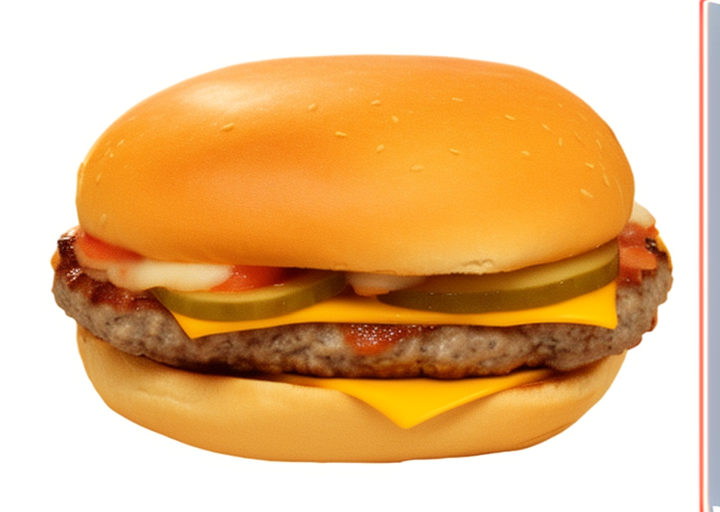}
        \caption{\texttt{cap-edit}}
    \end{subfigure}
    \hfill 
    \begin{subfigure}[b]{0.22\textwidth}
        \includegraphics[width=\textwidth]{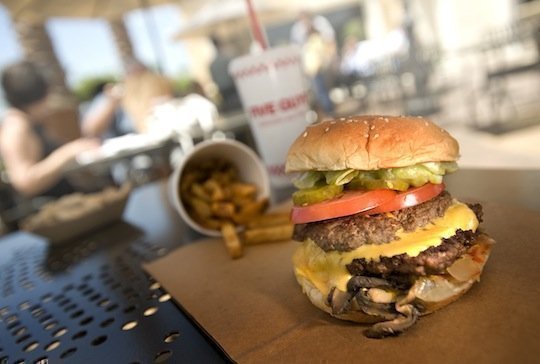}
        \caption{\texttt{cap-retrieve}}
    \end{subfigure}
    
    \caption{\emph{Source:} USA; \emph{Target:} USA --- \emph{BLIP caption}: a hamburger with cheese and pickles on a white background; \emph{LLM-edited caption}: 
a cheeseburger with pickles on a white bun. \texttt{e2e-instruct} heavily inserts flag colors, in \texttt{cap-edit} the LLM makes the bun white, \texttt{cap-retrieve} works well. Ideally, we do not want any change to be made in this case.}
\label{fig:usa-flag}
\end{figure*}

\begin{figure*}[ht!]
    \centering
    \begin{subfigure}[b]{0.22\textwidth}
        \includegraphics[width=\textwidth]{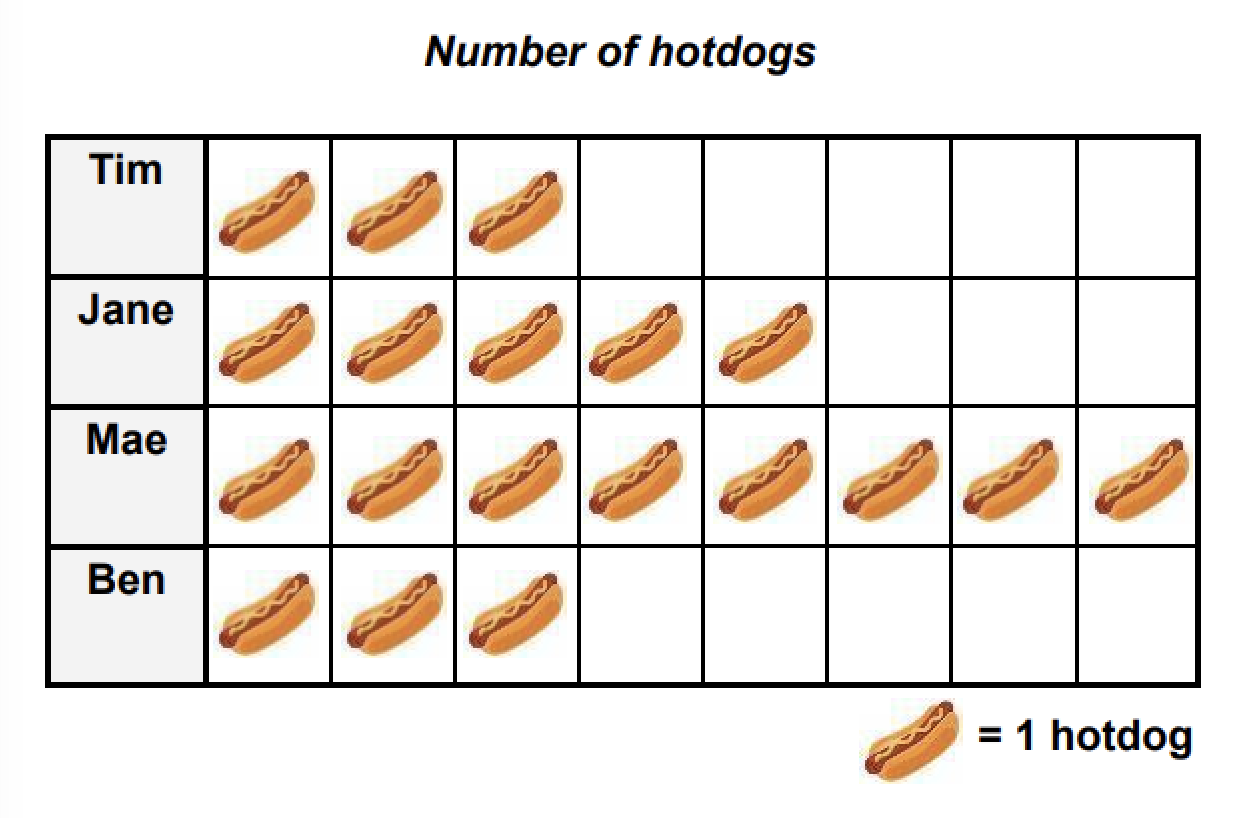}
        \caption{original}
    \end{subfigure}
    \hfill 
    \begin{subfigure}[b]{0.22\textwidth}
        \includegraphics[width=\textwidth]{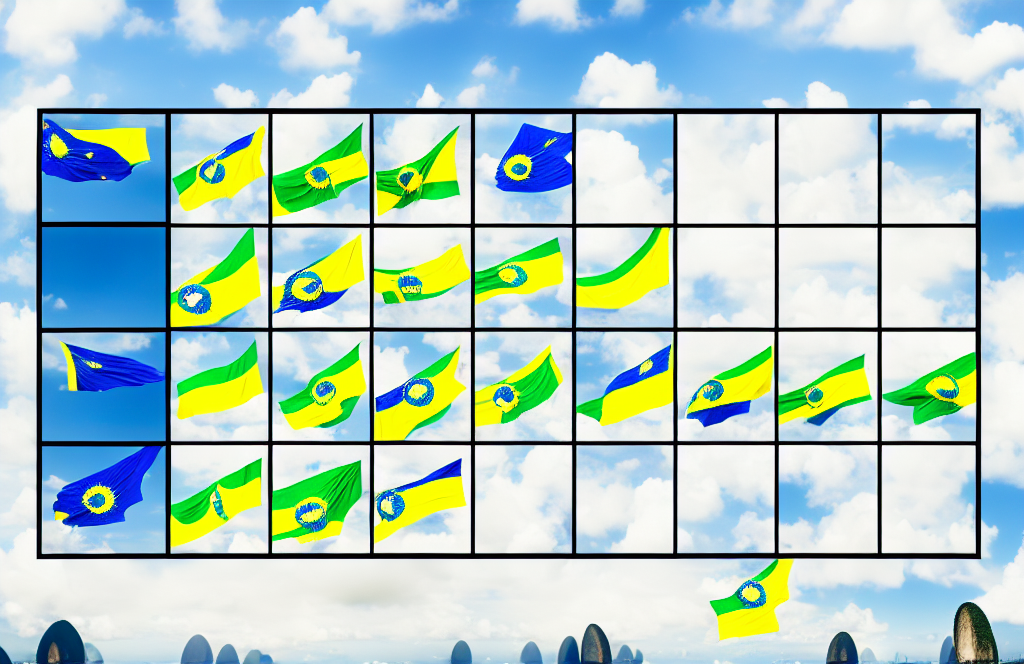}
        \caption{\texttt{e2e-instruct}}
    \end{subfigure}
    \hfill 
    \begin{subfigure}[b]{0.22\textwidth}
        \includegraphics[width=\textwidth]{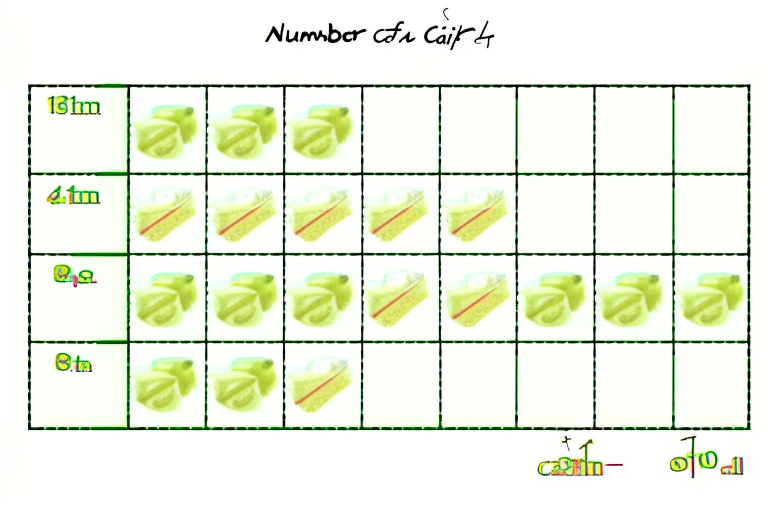}
        \caption{\texttt{cap-edit}}
    \end{subfigure}
    \hfill 
    \begin{subfigure}[b]{0.22\textwidth}
        \includegraphics[width=\textwidth]{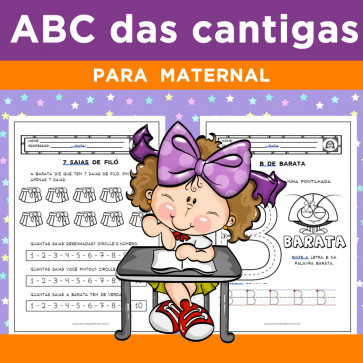}
        \caption{\texttt{cap-retrieve}}
    \end{subfigure}
    \caption{\emph{Application:} Story; \emph{Target:} Brazil --- \emph{Task}: Count the number of hotdogs. Here, we see a strong tendency to output elements of the map and flag colors in these models.}
    \label{fig:brazil-flag}
\end{figure*}

The models seem to associate flags and colors in them with a particular country/culture and includes these features in the edited images irrespective of the objects mentioned in the caption prompts.  Some examples can be seen in Figure \ref{fig:usa-flag}, where the American flag colors are applied over the Burger to make it relevant to the United States.  Similarly, Figure \ref{fig:brazil-flag} includes Brazil map and flag as part of the editing process. The code to run this pipeline is here.\footnote{\url{https://anonymous.4open.science/r/image-translation-6980/src/pipelines/e2e-instruct.py}} We simply pass in the original image with the instruction \emph{make this image culturally relevant to \texttt{COUNTRY}}.

\subsection{\texttt{cap-edit}: Caption, Text-edit, Image-edit}
\label{app:cap-edit} 
Spatial dimensions are highly preserved in this pipeline as can be seen in Figure \ref{fig:q5_part1}. This can sometimes lead to undesirable outcomes or the outputs to look unnatural, as shown below. The code to run this pipeline can be found here.\footnote{\url{https://anonymous.4open.science/r/image-translation-6980/src/pipelines/caption-llm_edit.py}}. The exact prompts used for captioning and LLM editing can be found in Table \ref{tab:prompts}. 

\begin{figure}[ht!]
    \centering
    \begin{subfigure}[b]{0.22\textwidth}
        \includegraphics[width=\textwidth]{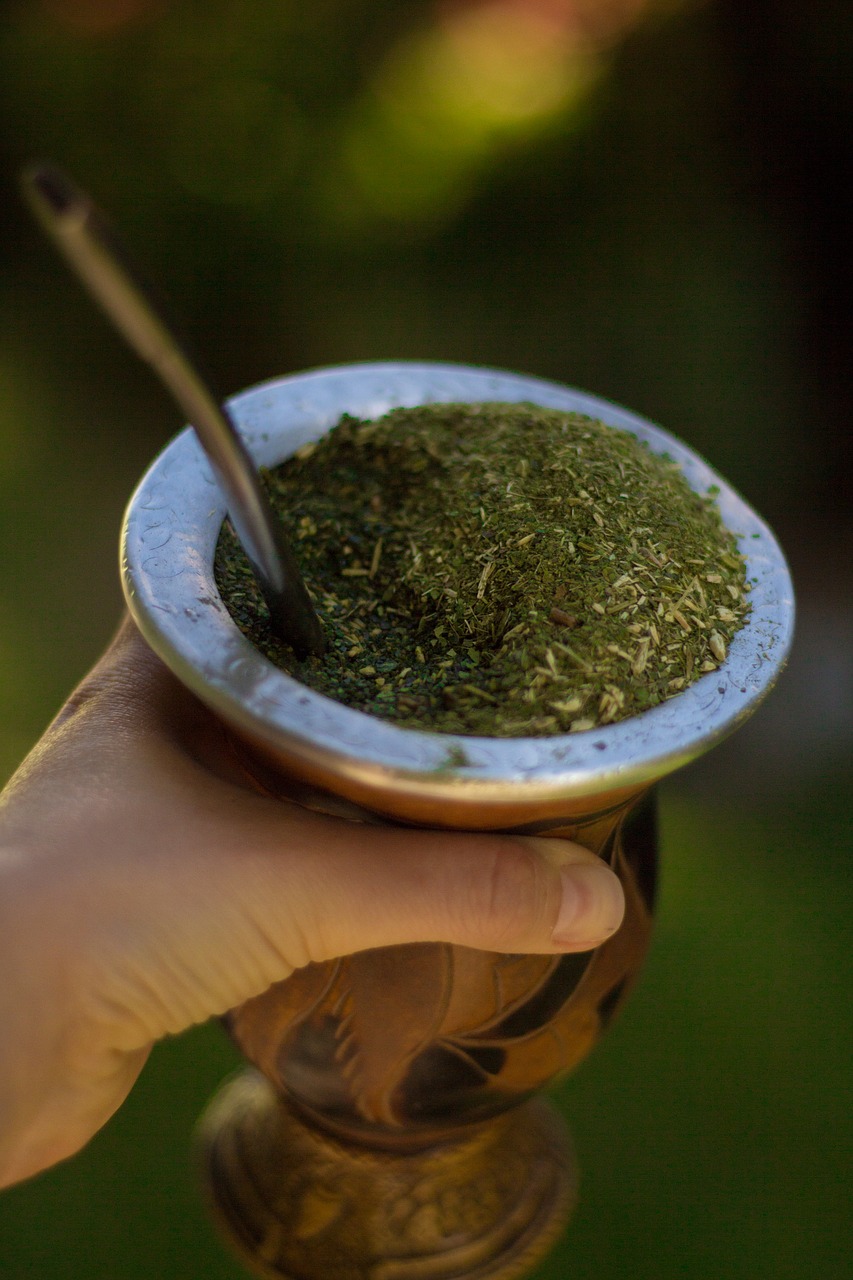}
        \caption{original}
    \end{subfigure}
    \hfill 
    \begin{subfigure}[b]{0.22\textwidth}
        \includegraphics[width=\textwidth]{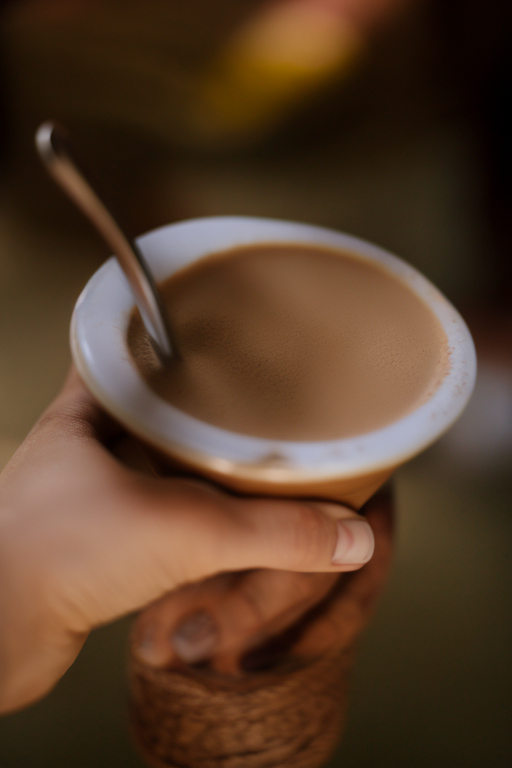}
        \caption{\texttt{cap-edit}}
    \end{subfigure}
    
    \caption{Example of how preserving the spatial layout of the original image can lead to unnatural looking outputs. Here, the final image shows \emph{a cup of chai}, but a typical cup of chai looks different in India.}
    \label{fig:chai}
\end{figure}

\subsection{\texttt{cap-retrieve}: Caption, Text-edit, Retrieval}
\label{app:cap-retrieve}
The obtained images through the retrieval pipeline seem to be noisy with a low precision but high recall. Some of the images are better representatives of that country's culture compared to the other two pipelines, given that they are real images. However, this pipeline also suffers from failure cases of retrieving images which may be too different from the source image or retrieving irrelevant outputs. Examples are shown below: 

\begin{figure}[ht!]
    \centering
    \begin{subfigure}[b]{0.22\textwidth}
        \includegraphics[width=\textwidth]{sections/figures/yerba.jpeg}
        \caption{original}
    \end{subfigure}
    \hfill 
    \begin{subfigure}[b]{0.22\textwidth}
        \includegraphics[width=\textwidth]{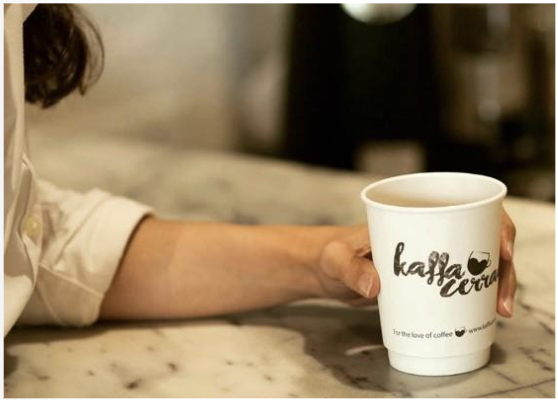}
        \caption{\texttt{cap-retrieve}}
    \end{subfigure}
    
    \caption{Example of how the retrieved output may at times look completely different from the original image.}
\end{figure}

\begin{figure}[ht!]
    \centering
    \begin{subfigure}[b]{0.22\textwidth}
        \includegraphics[width=\textwidth]{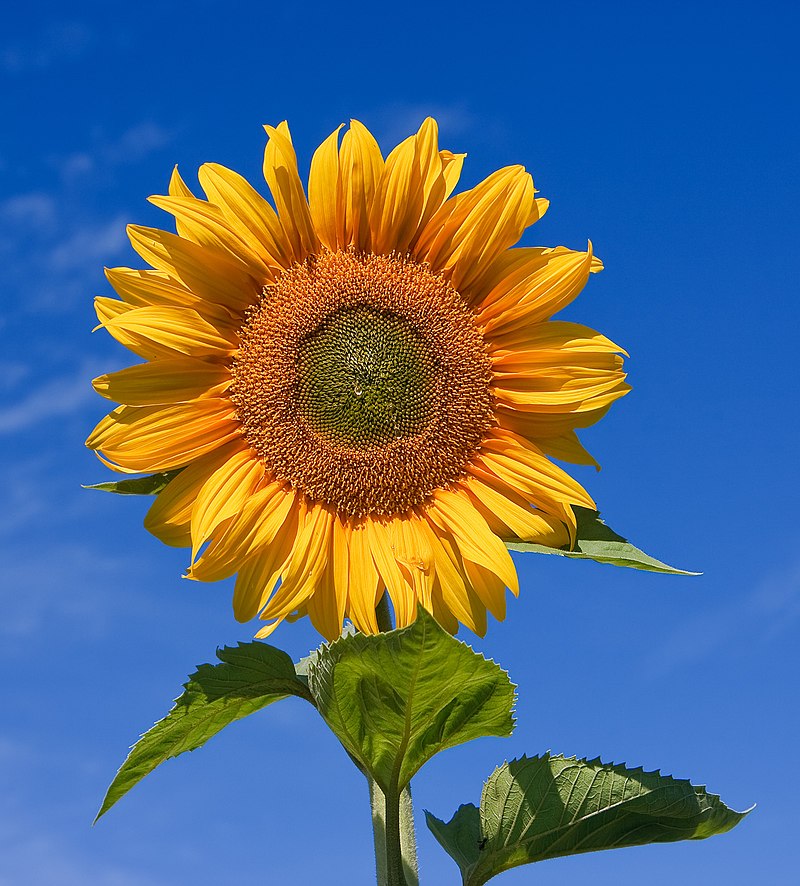}
        \caption{original}
    \end{subfigure}
    \hfill 
    \begin{subfigure}[b]{0.22\textwidth}
        \includegraphics[width=\textwidth]{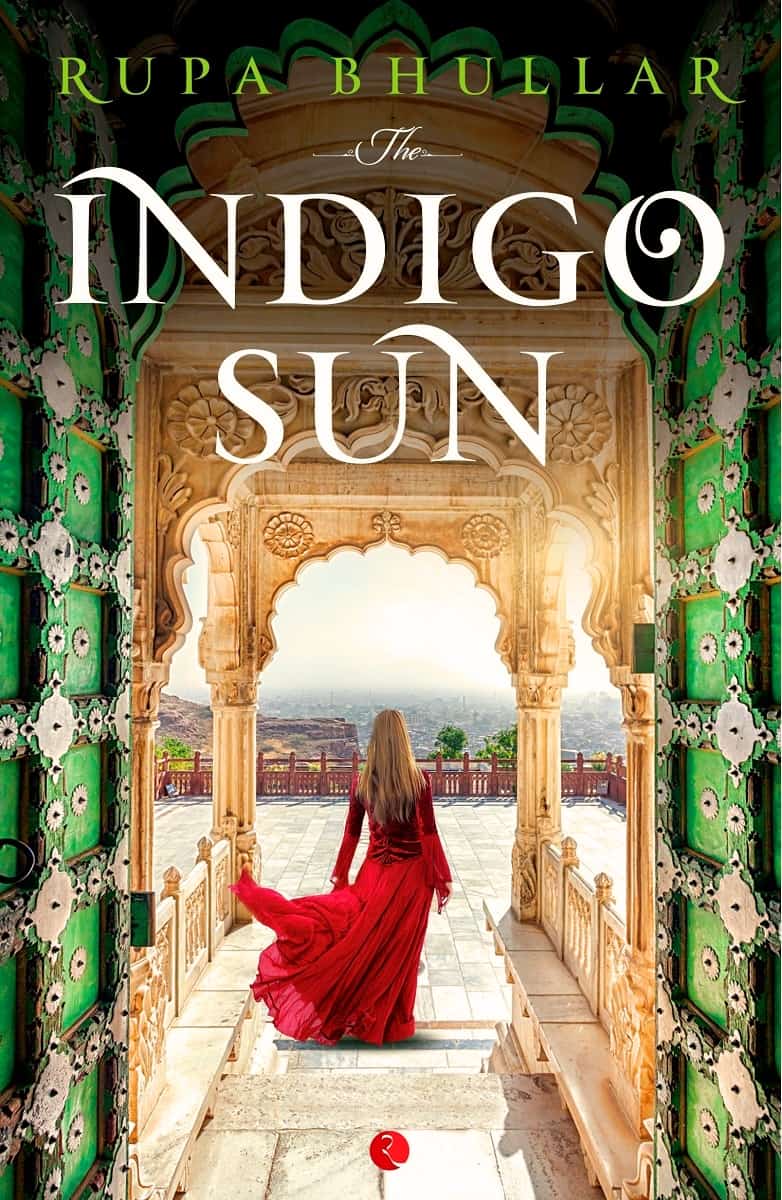}
        \caption{\texttt{cap-retrieve}}
    \end{subfigure}
    
    \caption{Example of how the retrieved output may be irrelevant/noisy. Here, we can see it behaving like a bag-of-words since the llm-edit used to prompt for retrieval is: \emph{A sunflower stands tall against the backdrop of a clear blue sky in India}.}
\end{figure}

\subsection{GPT4-o + GPT-4 + DALLE-3}
\label{app:cap-gen}
\begin{figure*}
    \centering
    \includegraphics[width=0.98\linewidth]{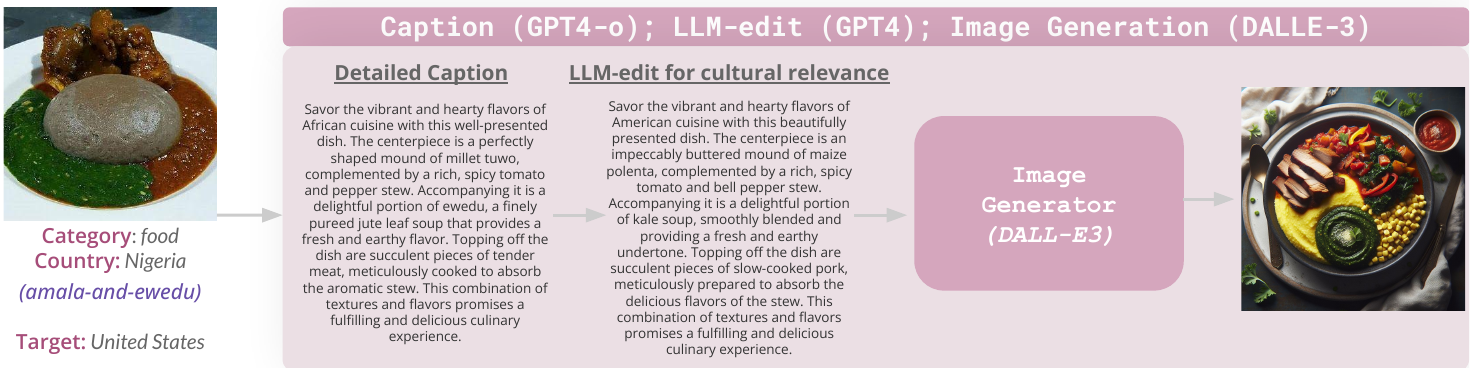}
    \caption{Pipeline for GPT4-based experiments. }
    \label{fig:dalle3}
\end{figure*}

We use the GPT-4 family of models for this pipeline. Since DALLE-3 works with detailed prompts \cite{betker2023improving}, we prompt GPT4-o to give detailed captions for images. We use GPT4 to edit these captions and prompt DALLE-3 to generate images. To make the images look natural, we add "photo, photograph, raw photo, analog photo, 4k, fujifilm photograph" to the prompt.\footnote{\url{https://www.reddit.com/r/dalle/comments/1au10g6/generate_realistic_pictures_with_dalle/}} Even then, the images do have a distinct style. Qualitatively, we observe that the captions and caption-edits capture fine-grained details which shorter captions in the previous two pipelines cannot. The overall pipeline can be found in Figure \ref{fig:dalle3}. All visualizations can be found in the released code repository. Note that GPT4-o + DALLE-3 outputs could not be human evaluated since their APIs were released on May 13, 2024. Further, the images' distinct style defeats the purpose of randomizing pipeline outputs for human evaluation. 

\begin{figure}
    \centering
    \includegraphics[width=0.98\linewidth]{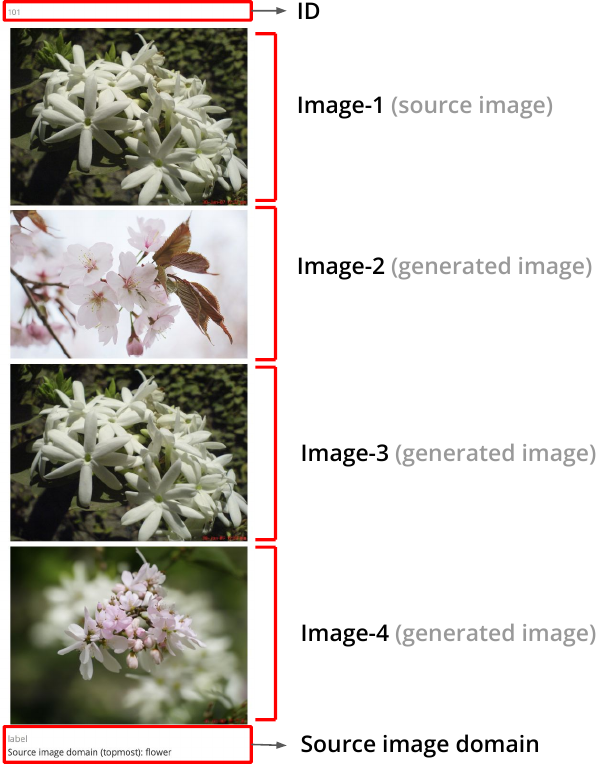}
    \caption{Screenshot of how one instance looks like for human evaluation on the Zeno platform. }
    \label{fig:screenshot}
\end{figure}

\section{Annotation Instructions}
\label{app:annot}
Our annotation and human evaluation instructions are as follows. We host our data on the Zeno\footnote{\url{https://zenoml.com/}} \cite{cabrera2023zeno} platform and hire people on Prolific\footnote{\url{https://www.prolific.com/}} to do the annotation and evaluation. Each worker is paid in the range of 10-15 USD per hour for the job. This work underwent IRB screening prior to conducting the evaluation.

\subsection{Part-1: Concept Collection}
This task is part of a research study conducted by \emph{[name]} at \emph{[place]}. In this research, we aim to create AI models that can generate images that are appropriate for different target audiences, such as people who live in different countries.

You will be given a set of universal categories that cover a diverse range of objects and events. These categories include things like bird, food, clothing, celebrations etc. You have to give Wikipedia links for 5 salient concepts for each category, that are most prevalent in your country and culture, for each of these categories. 

The two key requirements are for the concepts to be: \textbf{a)} commonly seen or representative of the speaking population of your country; \textbf{b)} ideally, to be physical and concrete. 

You have to make sure that the concept you select can be represented visually, i.e., an image can be used to represent the concept. 

A few examples for the food category for United States are given below:
\begin{itemize}
    \item \url{https://en.wikipedia.org/wiki/Hamburger}
    \vspace{-3mm}
    \item \url{https://en.wikipedia.org/wiki/Hot_dog}
\end{itemize}

\noindent \textbf{Note:} Links to wikipedia pages in English is preferred, but you can even provide a link to other languages if the concept is not present on English Wikipedia.

The categories are as follows: Bird, Mammal, Food, Beverages, Clothing, Houses, Flower, Fruit, Vegetable, Agriculture, Utensil/Tool, Sport, Celebrations, Education, Music, Visual Arts, Religion.

\subsection{Part-1: Image Collection}
This task is part of a research study conducted by \emph{[name]} at \emph{[place]}. In this research, we aim to create AI models that can generate images that are appropriate for different target audiences, such as people who live in different countries.

You will be given a set of universal categories that cover a diverse range of objects and events. These categories include things like bird, food, clothing, celebrations etc. You will also be given 5 concepts in each category that are highly relevant to your culture. 

Your task is to give us an image for each concept such that it reflects how it appears in your culture and native surroundings. Ideally this can be a wikipedia or wikimedia image itself. However, if you feel the wikipedia image is not appropriate, please provide us with a CC-licensed image from google image search. To filter for CC-licensing, look at the screenshot below.

A few examples for the food category for United States are given below:

1. \textbf{Concept (given to you)}: Hamburger (\url{https://en.wikipedia.org/wiki/Hamburger}). \\
\textbf{Image link (you have to provide)}:  \url{https://upload.wikimedia.org/wikipedia/commons/c/ce/McDonald%27s_Quarter_Pounder_with_Cheese%2C_United_States.jpg} \\

2. \textbf{Concept (given to you)}: Hotdog (\url{https://en.wikipedia.org/wiki/Hot_dog})\\
\textbf{Image link (you have to provide)}:
\url{https://upload.wikimedia.org/wikipedia/commons/thumb/b/b1/Hot_dog_with_mustard.png/220px-Hot_dog_with_mustard.png} 

Ensure that the images are clear and provide a good representation of the concept as it is experienced or seen in your culture and surroundings.

\subsection{Human Evaluation}

This task is part of a research study conducted by \emph{[name]} at \emph{[place]}. In this research, we aim to create AI models that can generate images that are appropriate for different target audiences, such as people who live in different countries. You need to be native to one of the following countries, and aware of its culture, to complete the task: 
Brazil, India, Japan, Nigeria, Portugal, Turkey, United States.

In this evaluation, you will be shown 4 images, as shown in the Figure \ref{fig:screenshot}. The top-most image (\emph{Image-1}) is sourced from the internet, from a diverse set of domains like agriculture, food, birds, education etc. This image is being edited to make it culturally relevant to your country and culture, using three state-of-the-art generative AI technologies (\emph{Image-2, Image-3, Image-4}). 

You will be asked whether you agree with six questions or statements about each of the images, from \textbf{5 (strongly agree)} to \textbf{1 (strongly disagree)}:  




\textbf{C0)} There are visual changes in the generated image, when compared with the source (top-most) image (\emph{1 → no visual change; 5 → high visual changes}). 

\textbf{C1)} The image contains similar content as the source image. For example, if the source is a food item, the target must also be a food item. Use the label to see which domain the source image is from (\emph{1 → dissimilar category; 5 → same category}). 

\textbf{C2)} The image maintains the spatial layout of the source image (this can be thought in terms of shapes and overall structure and placement of objects etc.) (\emph{1 → different layout; 5 → same layout}). 

\textbf{C3)} The image seems like it came from your country or is representative of your culture (\emph{1 → not culturally relevant; 5 → culturally relevant}).

\textbf{C4)} The image reflects naturally occurring scenes/objects (it does not look unnaturally edited and is something you can expect to see in the real world) (\emph{1 → unnatural; 5 → natural}). 

\textbf{C5)} This image is offensive to you, or is likely offensive to someone from your culture (\emph{1 → not offensive; 5 → offensive}). \\

\noindent \textbf{Stories} \\
\textbf{S1)} The image would match the text of the story in a children's storybook, as shown in the label. 

\noindent \textbf{S2)} The image seems like it came from your country or is representative of your culture. \\

\noindent \textbf{Education} \\
\textbf{E1)} The image can be used to teach the concept of the original worksheet, as shown in the label. 

\noindent \textbf{E2)} The image seems like it came from your country or is representative of your culture. \\

\noindent \textbf{[Optional]}: We would appreciate if you can share observations of certain patterns you found while doing the evaluation, post the study. For example, a few things we noticed are as follows:

1. Some models insert the flag or flag colors in the image, without any context, to increase the cultural relevance of it.

2. Some models exhibit color biases, like making things red/black, when asked to edit an image to make it culturally relevant to Japan.

3. Some models start inserting culturally prominent objects to increase relevance. For example, they commonly insert Mt. Fuji peaks, or cherry blossoms, to make an image culturally relevant to Japan.

\subsection{Observations as noted by human evaluators}
This is the feedback received for the optional comments in the human evaluation as asked for above. Almost everyone found outputs to be semantically incoherent with random insertions of colors, cultural entities, flag elements and so on, uncovering several biases and gaps that these models have today.  

\subsubsection{Brazil}

\begin{itemize}
    \item \emph{Overall, I noticed that the colors of Brazil’s flag were extensively used in various contexts, creating an unnatural effect on the subject of the pictures. I cannot precisely articulate why, but I felt that these images gave me an impression of Africa rather than Brazil, even though Brazil is an extremely diverse country with a significant African influence. Additionally, I observed numerous abstract representations where only the basic shape from the original picture was retained.}
    \item \emph{Some images had the colors of the Brazilian flag as if "superimposed" on the objects and images, without making sense with the figure itself}
\end{itemize}

\subsubsection{Japan}

\begin{itemize}
    \item \emph{There are not enough variations to represent Japan. Commonly used subjects - cherry blossoms, pine trees, Mt.Fuji}
    \item \emph{Characters in Japanese children's picture books tend to have American-leaning faces, making Japanese faces look more adult-oriented}
    
\end{itemize}

\subsubsection{India}

\begin{itemize}
    \item \emph{Models have put some improper Indian images with only cultural costume and also found many bad generated faces}
\end{itemize}

\subsubsection{Nigeria}

\begin{itemize}
    \item \emph{Some models just changed the pictures to green in an attempt to make it look Nigerian. Images did not match the description.}
    \item \emph{Models has a lot of black scary images that did not fit the context and doesn't make it culturally relevant to Nigeria. Images generated did not match the original image neither was it relevant to the Nigerian culture.}
\end{itemize}

\subsubsection{Portugal}

\begin{itemize}
\item \emph{In the math worksheets, for so many times it was generated a picture that would add, random parts of the portugese flag or colors making no sense at all and sometimes it looks like Morocco}
\item \emph{Some problems are not related to mathematics: such as the question of associating what each "element" can carry on its back}

\end{itemize}

\subsubsection{Turkey}

\begin{itemize}
\item \emph{Observed that a lot of the edited images included turkey (the animal) illustrations, and also some of the edited images included Turkish flag, mosques, Turkish food, Turkish tea and some clothing styles that were mostly used in ancient times. Some of the edited images were only consisting of the colors of the Turkish flag, which are red and white.
}
\item \emph{In some instances, where there was a person of color or a person with a different ethnicity in the topmost image, the skin color of the person was changed in the edited images and sometimes beards were added on men, and headscarves were added on women}

\end{itemize}

\subsubsection{USA}

\begin{itemize}
    \item \emph{I did notice that in the majority of images with people/faces, that the AI image rearranged/disoriented the facial features}
    \item \emph{The AI images related to plants, food and nature seem to be more natural in the edits and effects and way more natural than when applying the same change of effects on people}
\end{itemize}

\begin{figure*}[ht!]
    \centering
    \begin{subfigure}[b]{0.22\textwidth}
        \includegraphics[width=\textwidth]{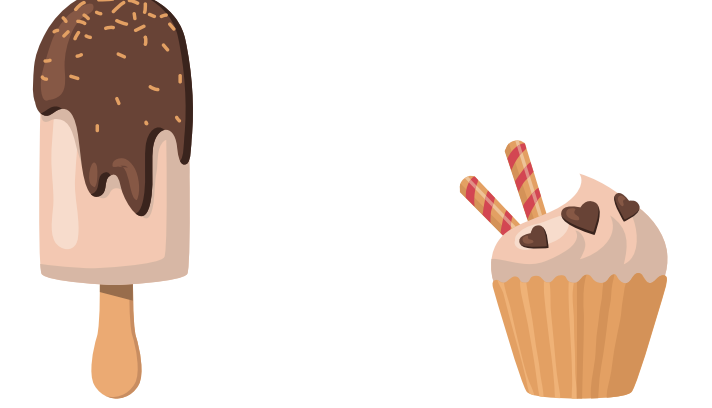}
        \caption{original}
    \end{subfigure}
    \hfill 
    \begin{subfigure}[b]{0.22\textwidth}
        \includegraphics[width=\textwidth]{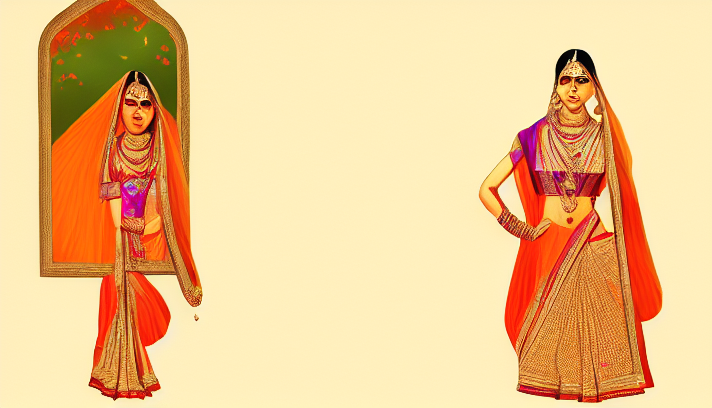}
        \caption{\texttt{e2e-instruct}}
    \end{subfigure}
    \hfill 
    \begin{subfigure}[b]{0.22\textwidth}
        \includegraphics[width=\textwidth]{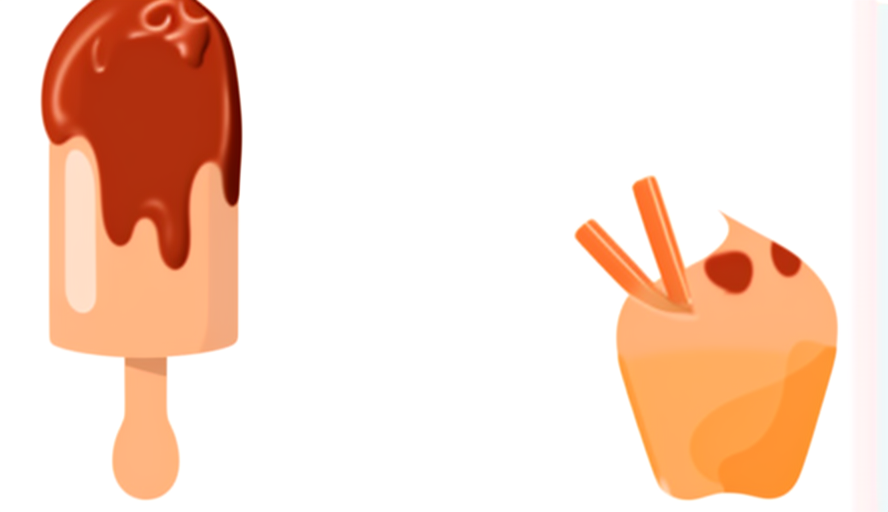}
        \caption{\texttt{cap-edit}}
    \end{subfigure}
    \hfill 
    \begin{subfigure}[b]{0.22\textwidth}
        \includegraphics[width=\textwidth]{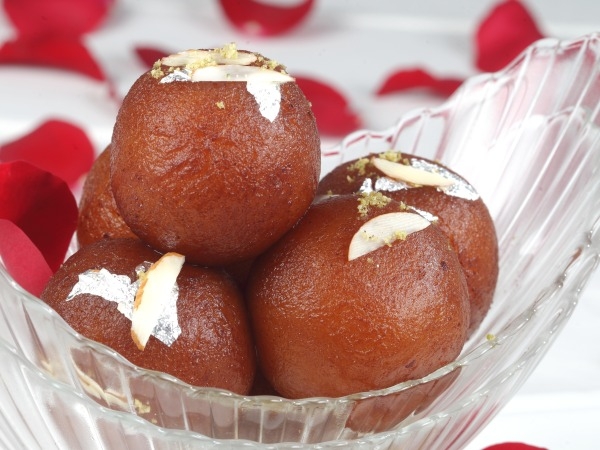}
        \caption{\texttt{cap-retrieve}}
    \end{subfigure}
    
    \caption{\emph{Application:} Education; \emph{Target:} India --- \emph{Task}: Pick the largest one among the two icecreams; \emph{InstructBLIP caption}: a cupcake and an ice cream pop on a white background; \emph{LLM-edited caption}: a gulab jamun and a kulfi on a white background. \texttt{e2e-instruct} inserts women in traditional indian clothing not relevant to the task, the LLM makes a pretty good edit but the image-editing model in \texttt{cap-edit} probably doesn't understand indian sweets like gulab jamun and kulfi, and the retriever in \texttt{cap-retrieve} only retrieves one item of two.}
\end{figure*}

\begin{figure*}[ht!]
    \centering
    \begin{subfigure}[b]{0.22\textwidth}
        \includegraphics[width=\textwidth]{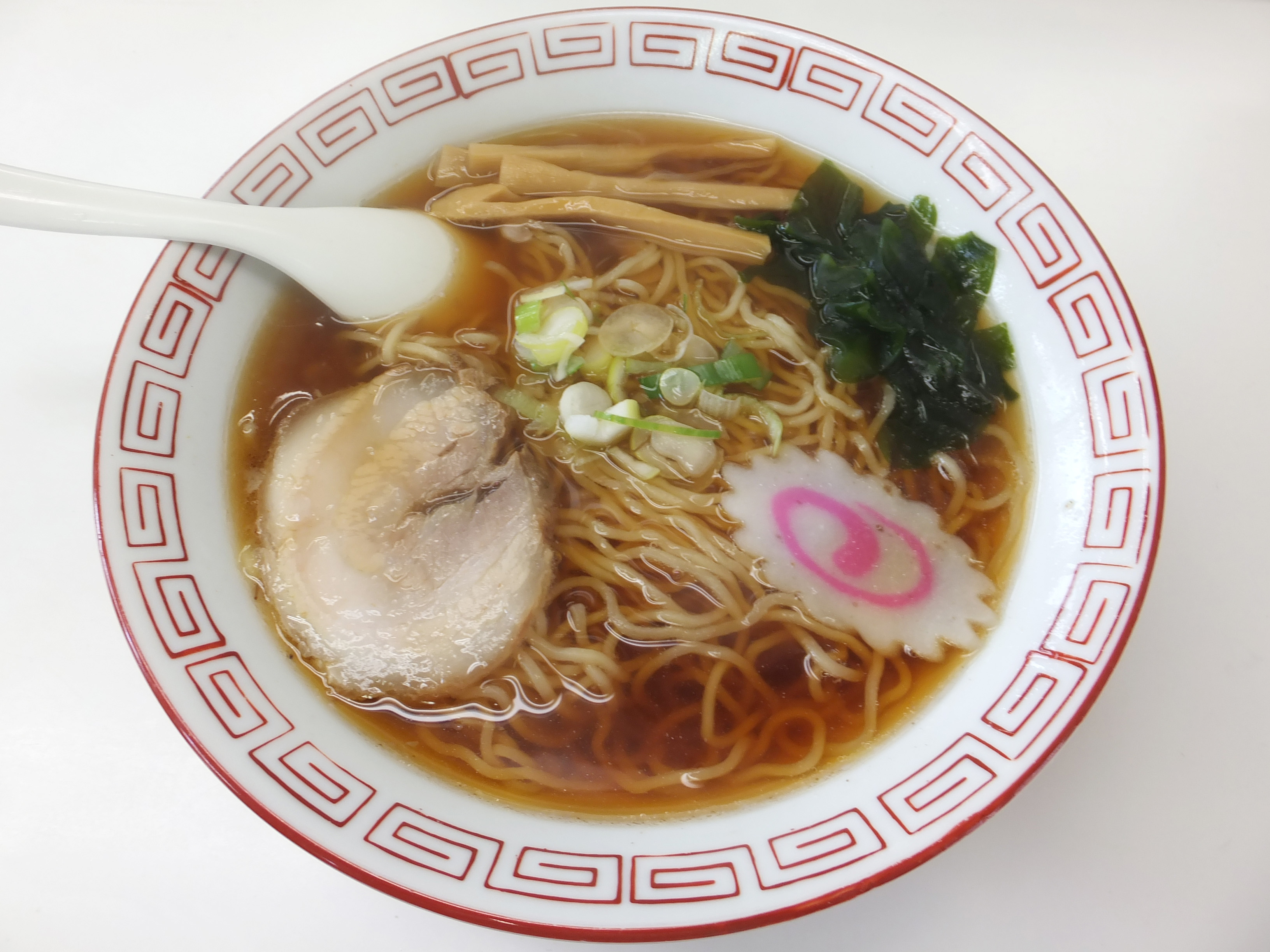}
        \caption{original}
    \end{subfigure}
    \hfill 
    \begin{subfigure}[b]{0.22\textwidth}
        \includegraphics[width=\textwidth]{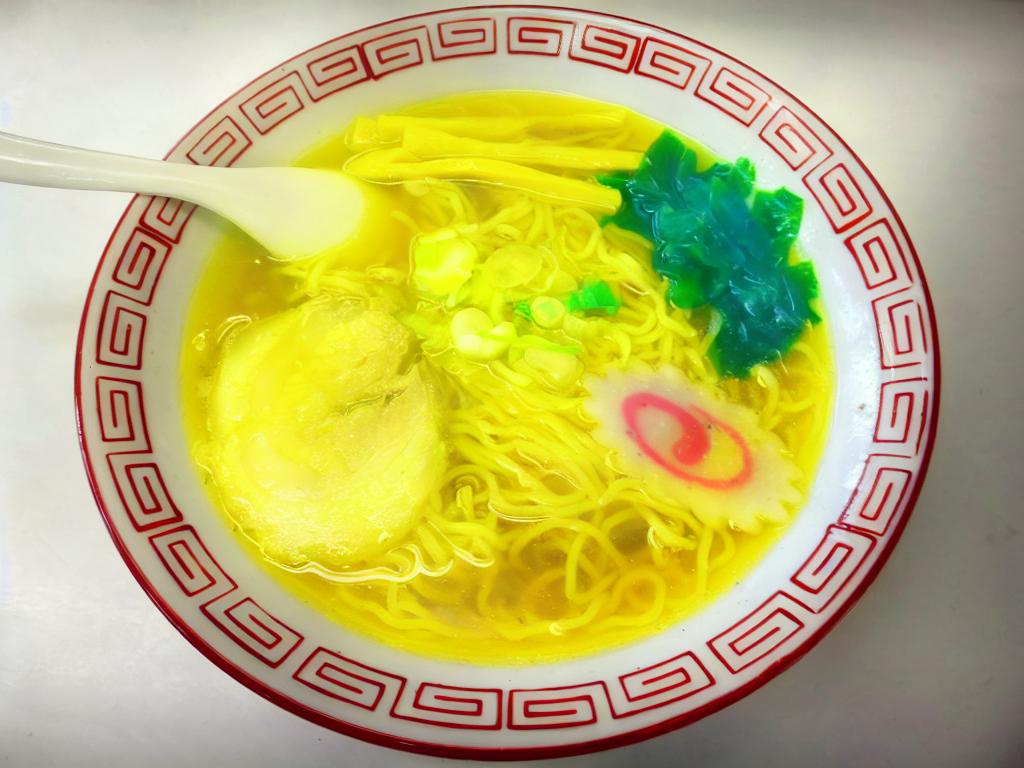}
        \caption{\texttt{e2e-instruct}}
    \end{subfigure}
    \hfill 
    \begin{subfigure}[b]{0.22\textwidth}
        \includegraphics[width=\textwidth]{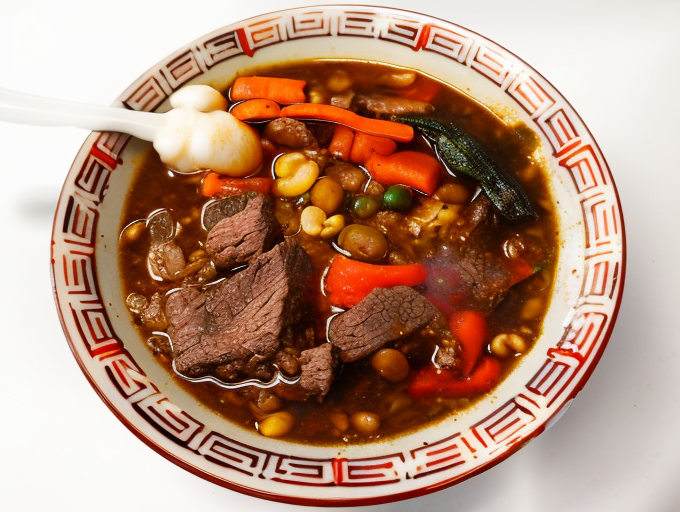}
        \caption{\texttt{cap-edit}}
    \end{subfigure}
    \hfill 
    \begin{subfigure}[b]{0.22\textwidth}
        \includegraphics[width=\textwidth]{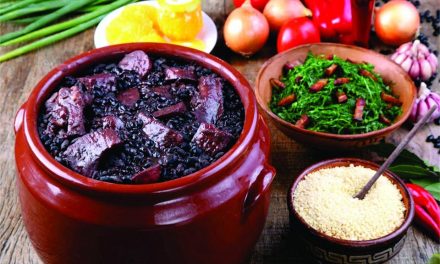}
        \caption{\texttt{cap-retrieve}}
    \end{subfigure}
    \caption{\emph{Source:} Japan; \emph{Target:} Brazil --- \emph{BLIP caption}: a bowl of ramen with meat and vegetables; \emph{LLM-edited caption}: 
a bowl of feijoada with beef and vegetables. \texttt{e2e-instruct} simply inserts flag colors, \texttt{cap-edit} highly preserves structural layout, \texttt{cap-retrieve} retrieves a natural image but is structurally different from the source.}

\end{figure*}

\begin{figure*}[ht!]
    \centering
    \begin{subfigure}[b]{0.22\textwidth}
        \includegraphics[width=\textwidth]{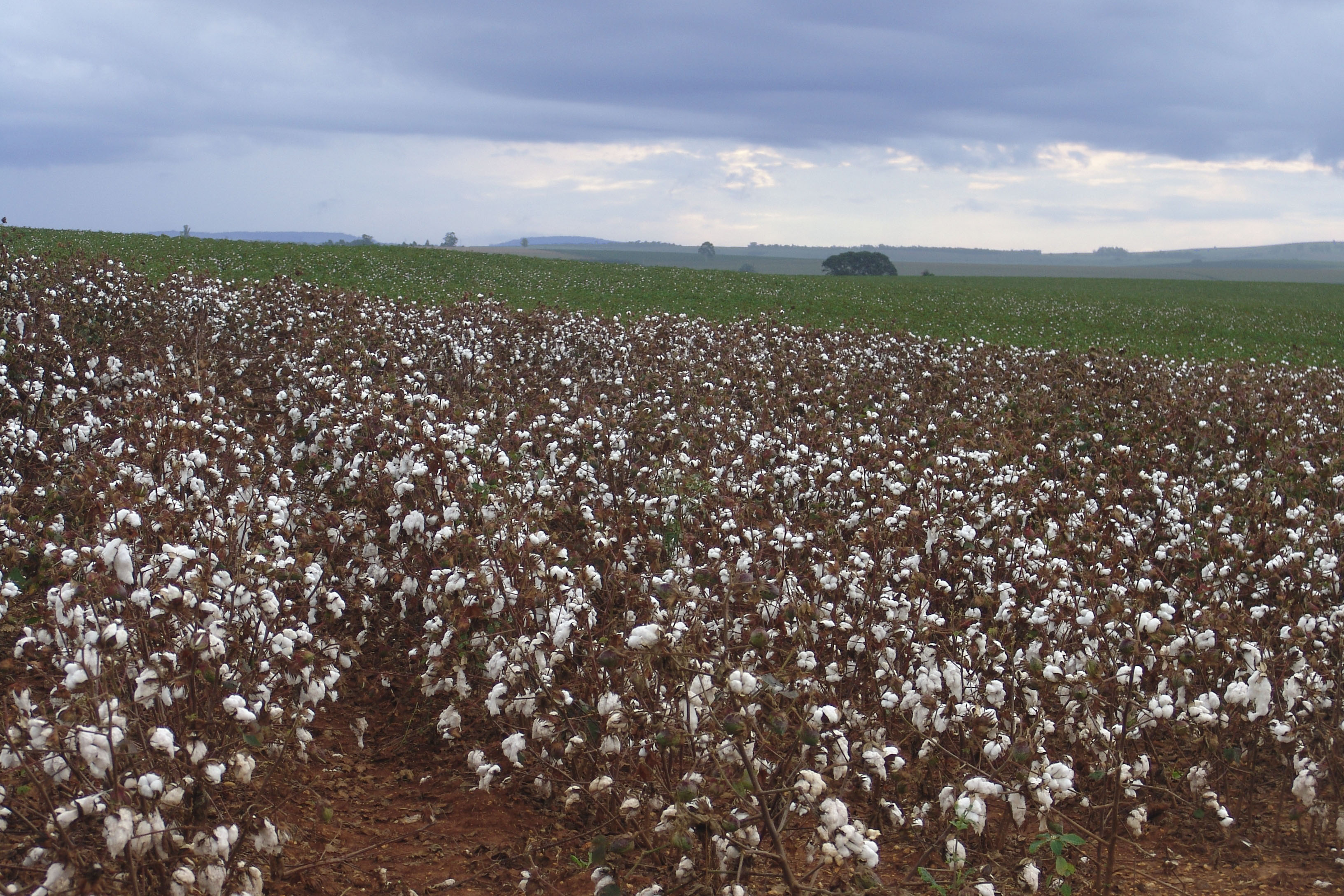}
        \caption{original}
    \end{subfigure}
    \hfill 
    \begin{subfigure}[b]{0.22\textwidth}
        \includegraphics[width=\textwidth]{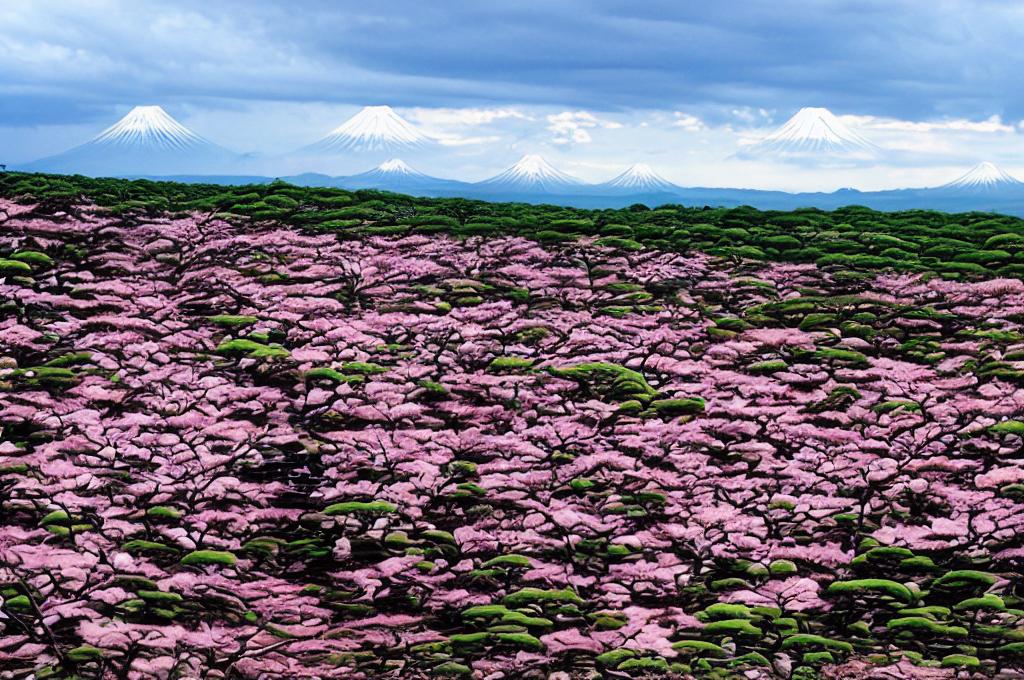}
        \caption{\texttt{e2e-instruct}}
    \end{subfigure}
    \hfill 
    \begin{subfigure}[b]{0.22\textwidth}
        \includegraphics[width=\textwidth]{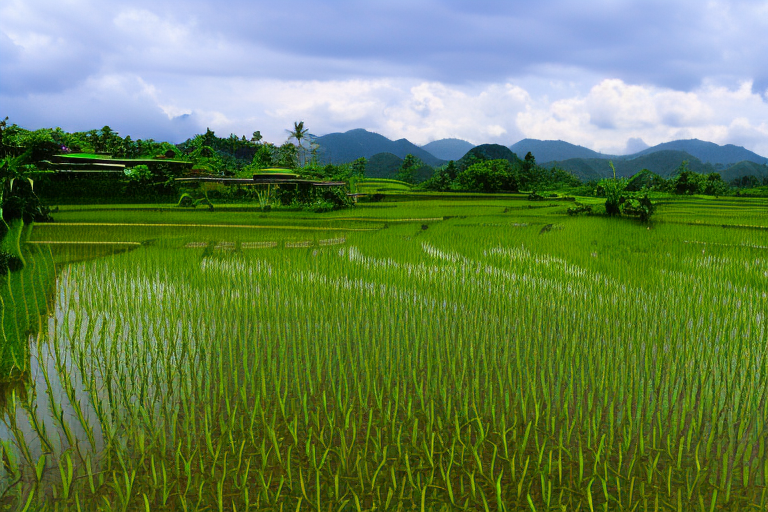}
        \caption{\texttt{cap-edit}}
    \end{subfigure}
    \hfill 
    \begin{subfigure}[b]{0.22\textwidth}
        \includegraphics[width=\textwidth]{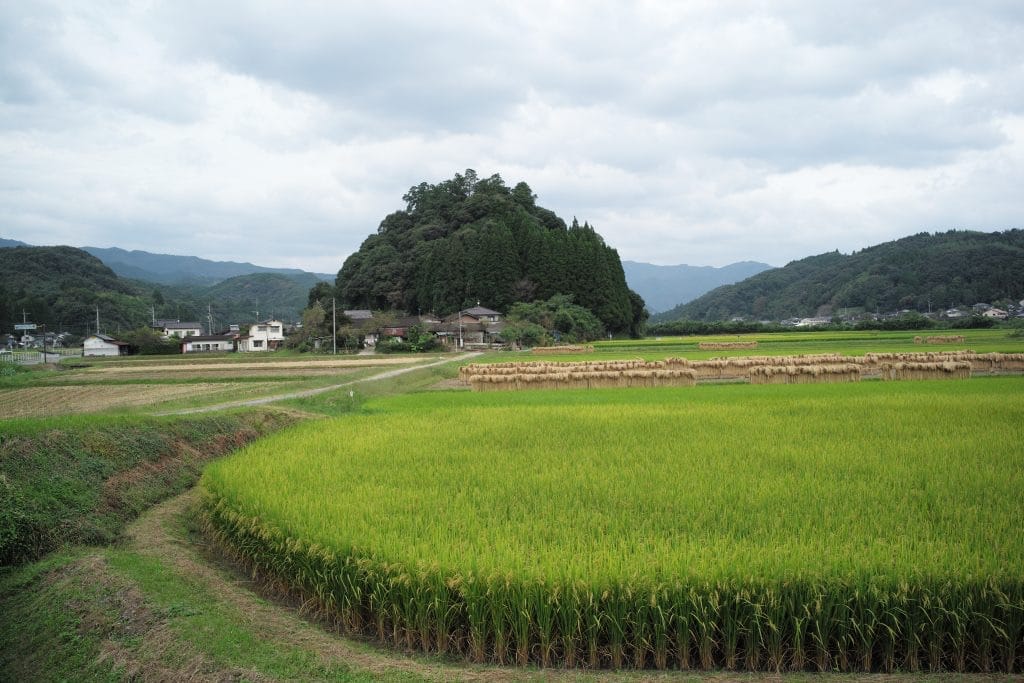}
        \caption{\texttt{cap-retrieve}}
    \end{subfigure}
    \caption{\emph{Source:} India; \emph{Target:} Japan --- \emph{BLIP caption}: a field of cotton plants; \emph{LLM-edited caption}: 
a rice paddy field. \texttt{e2e-instruct} inserts sakura blossoms and multiple Mt. Fuji peaks in the background, \texttt{cap-edit} highly preserves structural layout but looks pretty realistic here.}
\end{figure*}

\begin{figure*}[ht!]
    \centering
    \begin{subfigure}[b]{0.22\textwidth}
        \includegraphics[width=\textwidth]{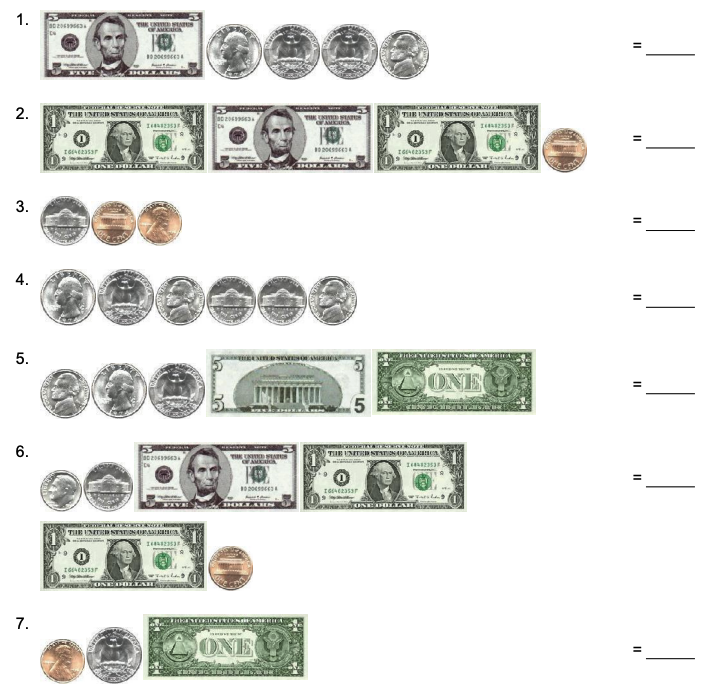}
        \caption{original}
    \end{subfigure}
    \hfill 
    \begin{subfigure}[b]{0.22\textwidth}
        \includegraphics[width=\textwidth]{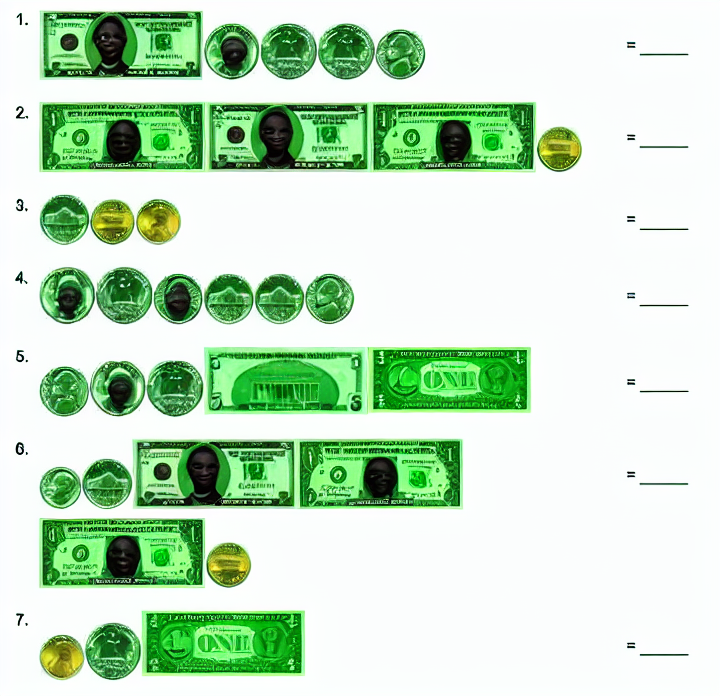}
        \caption{\texttt{e2e-instruct}}
    \end{subfigure}
    \hfill 
    \begin{subfigure}[b]{0.22\textwidth}
        \includegraphics[width=\textwidth]{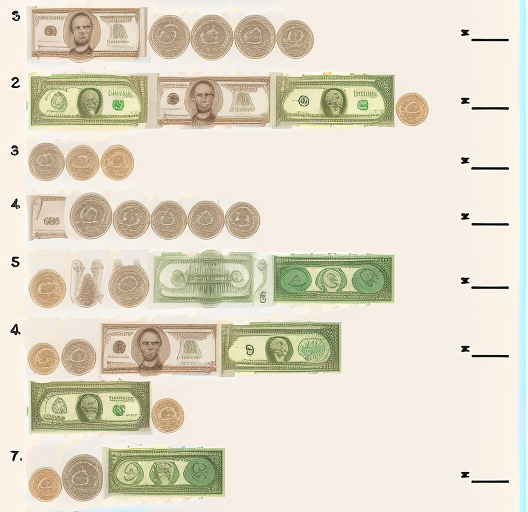}
        \caption{\texttt{cap-edit}}
    \end{subfigure}
    \hfill 
    \begin{subfigure}[b]{0.22\textwidth}
        \includegraphics[width=\textwidth]{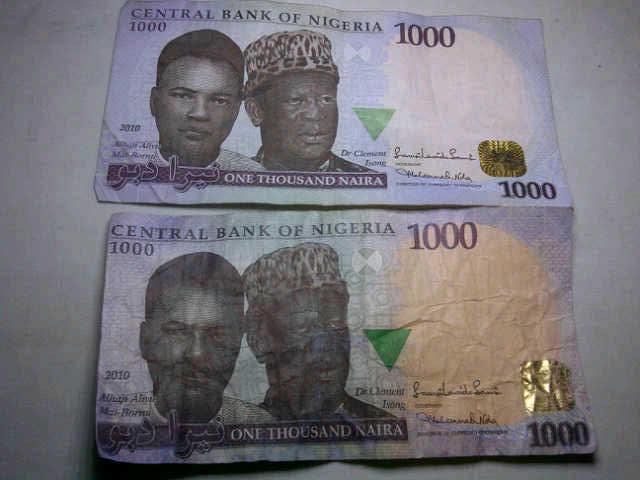}
        \caption{\texttt{cap-retrieve}}
    \end{subfigure}
    
    \caption{\emph{Application:} Education; \emph{Target:} Nigeria --- \emph{Task}: Add the US currency notes; \emph{InstructBLIP Caption}: a math worksheet with coins and notes on it \emph{LLM-edit Caption}: a math worksheet with Naira coins and notes on it. We see the pipelines exhibiting strong color bias both for the notes and the background itself.}
\end{figure*}

\begin{figure*}[ht!]
    \centering
    \begin{subfigure}[b]{0.22\textwidth}
        \includegraphics[width=\textwidth]{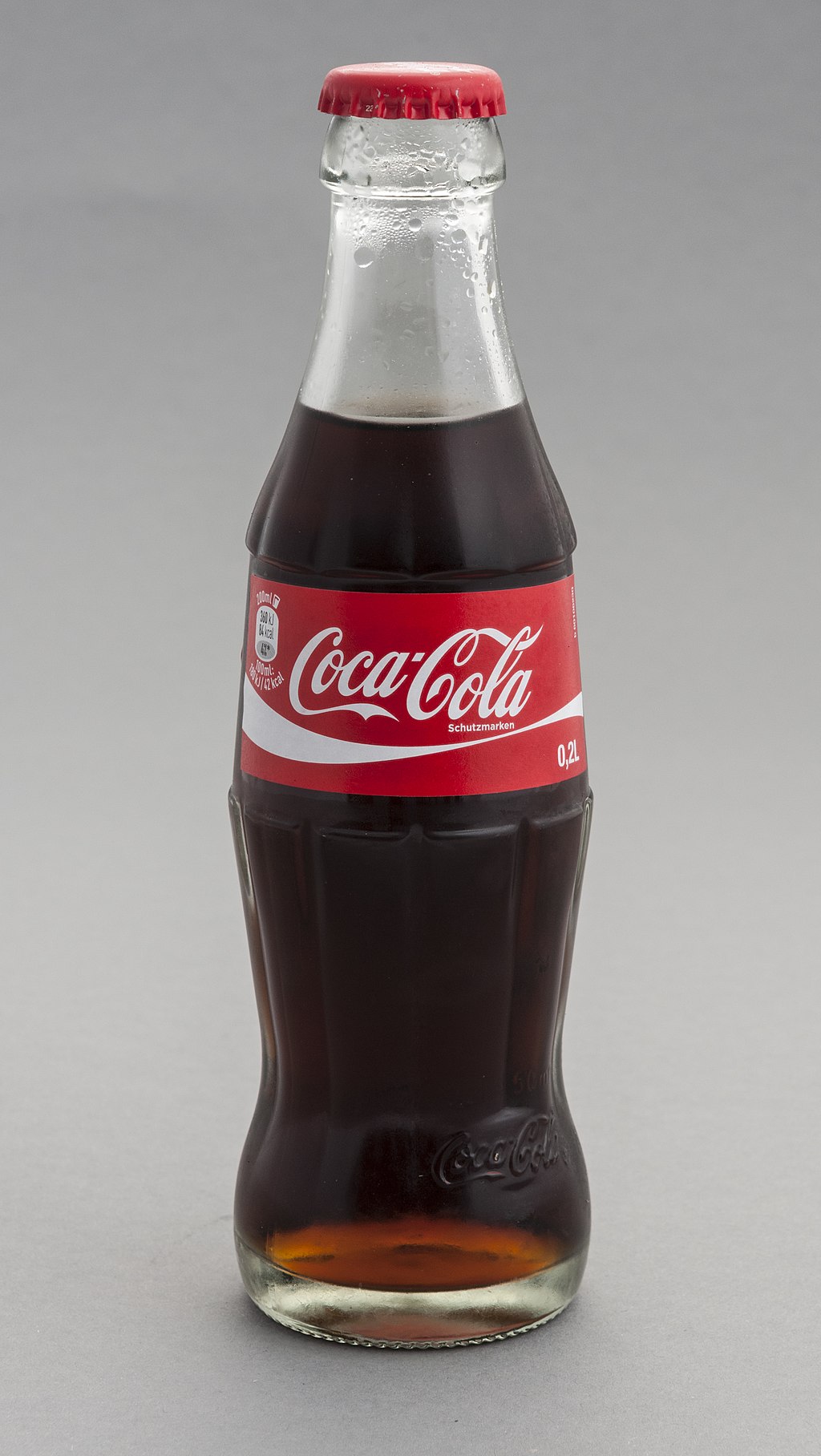}
        \caption{original}
    \end{subfigure}
    \hfill 
    \begin{subfigure}[b]{0.22\textwidth}
        \includegraphics[width=\textwidth]{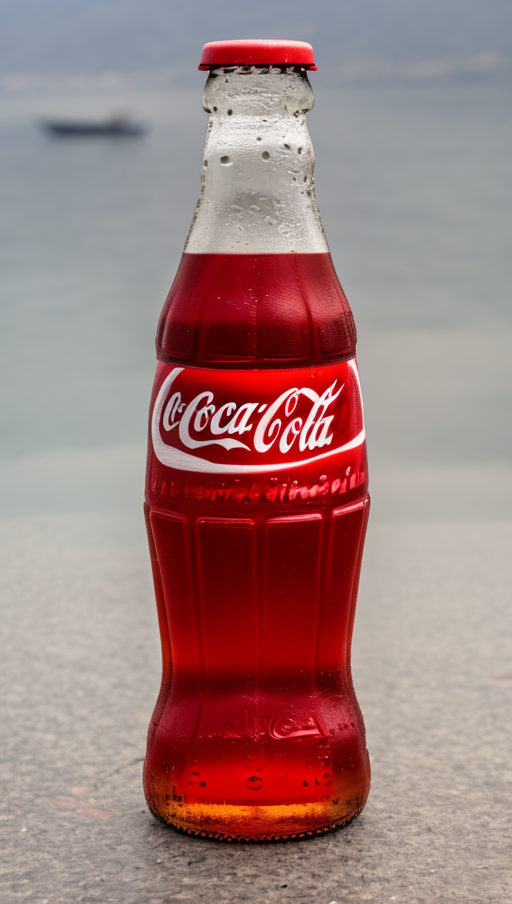}
        \caption{\texttt{e2e-instruct}}
    \end{subfigure}
    \hfill 
    \begin{subfigure}[b]{0.22\textwidth}
        \includegraphics[width=\textwidth]{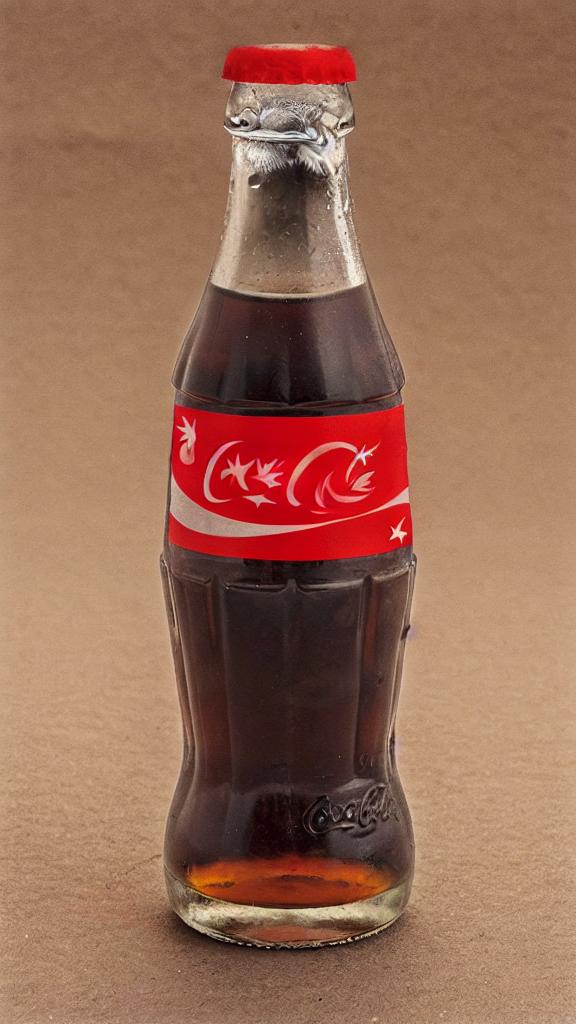}
        \caption{\texttt{cap-edit}}
    \end{subfigure}
    \hfill 
    \begin{subfigure}[b]{0.22\textwidth}
        \includegraphics[width=\textwidth]{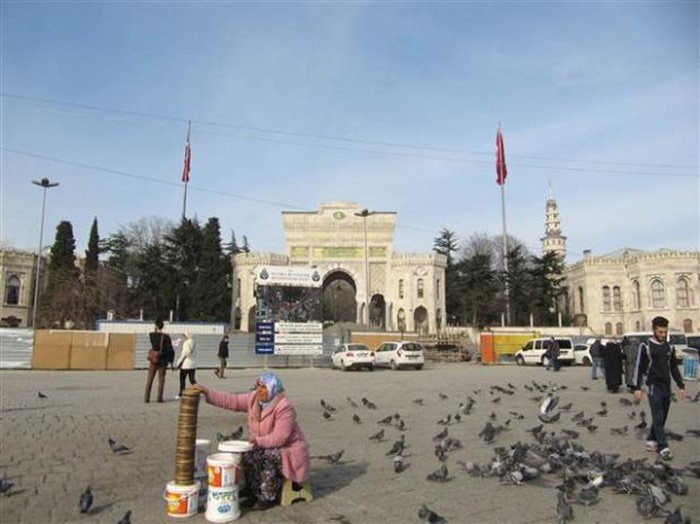}
        \caption{\texttt{cap-retrieve}}
    \end{subfigure}
    \caption{\emph{Source:} United States; \emph{Target:} Turkey --- \emph{BLIP caption}: a coca cola bottle with a red lid; \emph{LLM-edited caption}: 
a bottle of coca cola with a red cap in Turkey. \texttt{e2e-instruct} doesn't know that coca-cola is black, and makes it red for Turkey, \texttt{cap-edit} adds flag details to the logo and the LLM also simply adds "turkey" in the caption while \texttt{cap-retrieval} just produces an irrelevant output.}
\end{figure*}

    

\section{Quantitative metrics}
\label{app:quant_eval}
\begin{figure}
    \centering
    \includegraphics[width=\linewidth]{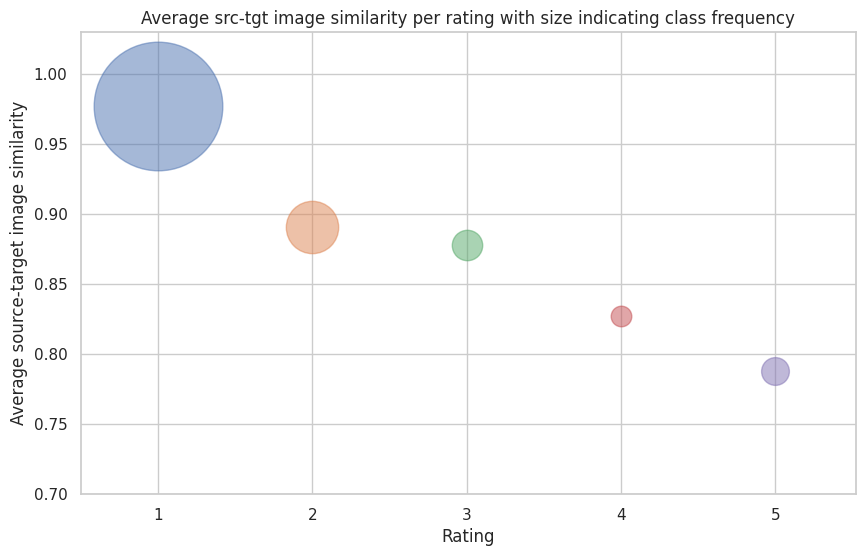}
    \caption{\texttt{correlation plot}, capturing linear correlation between human and machine evaluation for Brazil}
    \label{fig:correlation}
\end{figure}

We find a linear correlation between image-image similarity scores and human evaluation ratings on \textbf{C0:} \texttt{visual-change}. This helps us determine a threshold beyond which, on average, images get a visual-change score of 1 or 2 (1 means no visual change). A correlation plot for one of the countries is shown in Figure \ref{fig:correlation}. 

For the application-oriented evaluation, we simply ask whether the edited image can be used to solve the same task (in education) or whether it matches the title of the story (for stories). However, if the image is not edited at all, pipelines would still score high on this question, thus biasing our analysis. Since we notice a linear correlation in image-similarity and human ratings for the same question in \emph{concept} evaluation, we determine a threshhold in image similarity beyond which humans give a rating of 1 or 2 to the image (1 means no visual change). This threshhold typically hovers around 0.95-0.97 for each country.

\begin{figure*}
    \centering
    \includegraphics[width=\linewidth]{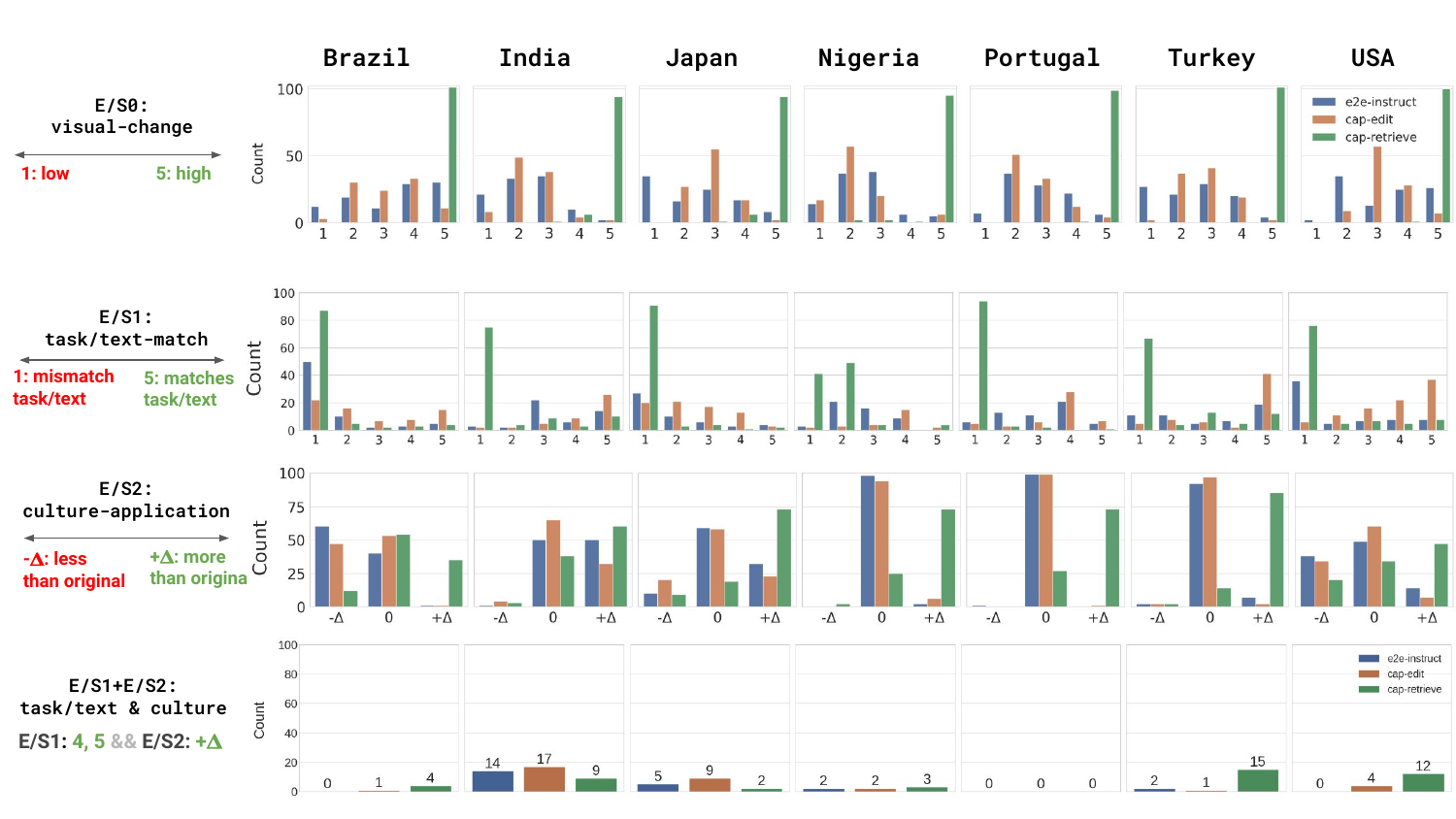}
    \caption{\emph{Human ratings for the application dataset}: Our goal is to test whether the edited image can be used for the application as before (\textbf{E/S1}), and whether it increases cultural relevance (\textbf{E/S2}). We plot the count of images that can do both above (\textbf{E/S1+E/S2}), and observe that even the best pipeline cannot transcreate any image successfully in some cases, like for Portugal.}
    \label{fig:app-results}
\end{figure*}

For E1 and S1 application plots in Figure \ref{fig:app-results}, we employ these thresholds to filter images that haven't been edited at all. Images whose image-similarity scores greater than the thresholds calculated are filtered out, ensuring that only those images that have been edited are considered for further analysis.


\begin{figure}
    \centering
    \includegraphics[width=\linewidth]{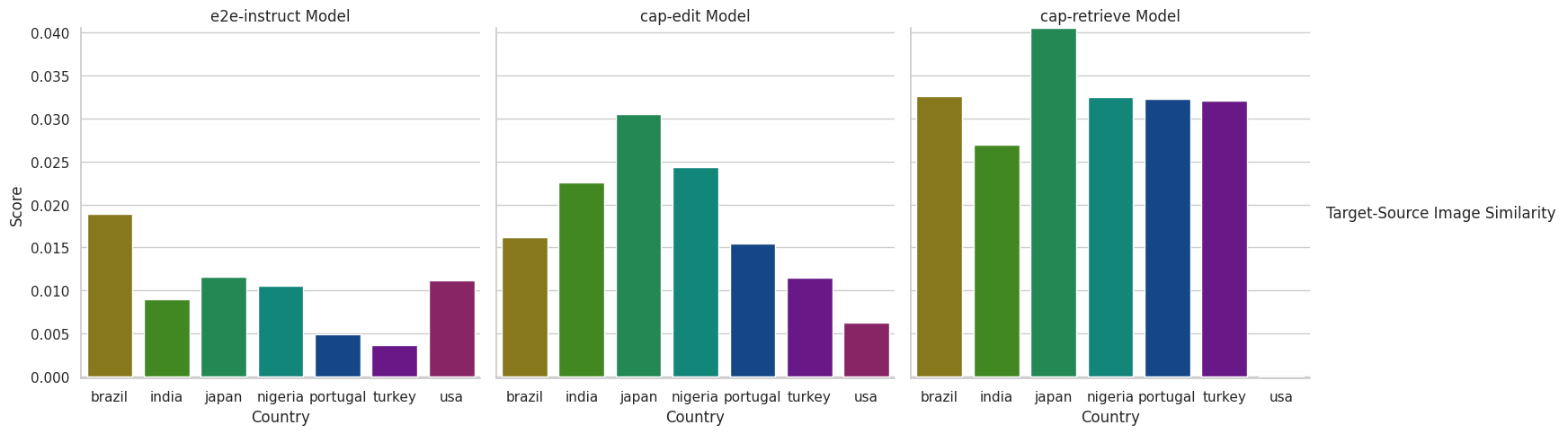}
    \caption{\texttt{target-source similarity}, capturing the difference in image-text similarity scores between target and source}
    \label{fig:target_source_sim}
\end{figure}

\begin{figure}
    \centering
    \includegraphics[width=\linewidth]{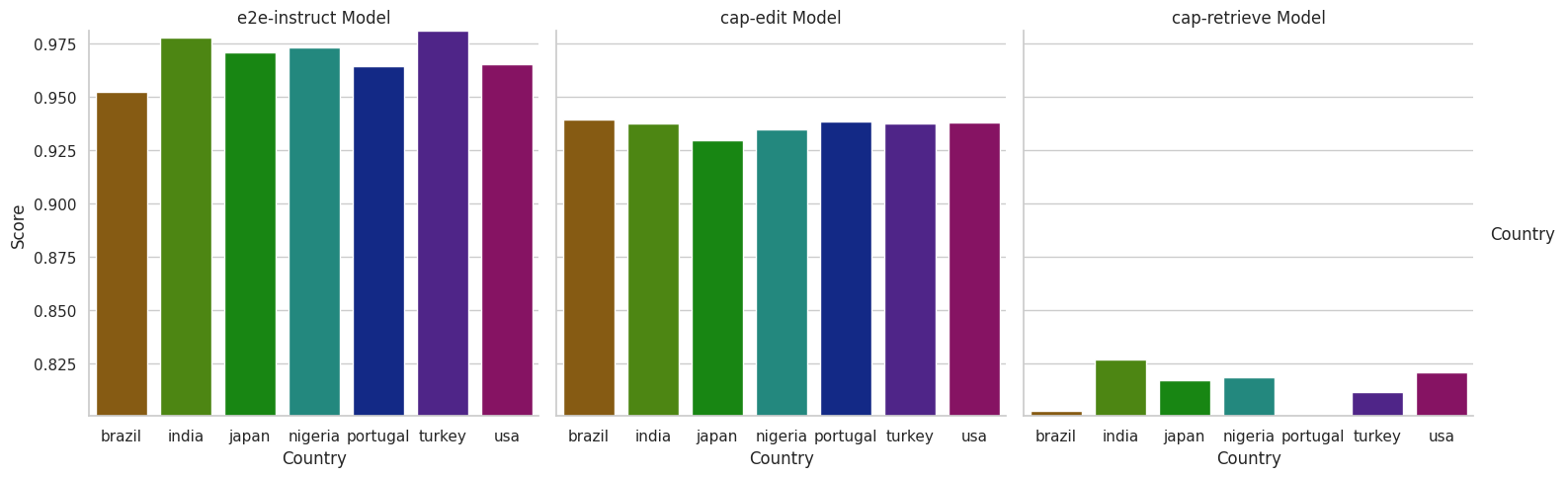}
    \caption{\texttt{image similarity difference}, capturing the difference in image similarity scores between target and source}
    \label{fig:image_sim}
\end{figure}

\section{Continued analysis of human evaluation}
\label{app:human_eval}
We continue analysis of questions asked in Table \ref{tab:qs_part1} below:

\textbf{C0:} \texttt{visual-change} -- First we ask whether the image has been edited at all, to help understand if the edits make sense in the questions that follow. Across all countries, \texttt{cap-retrieve} maximally edits images, with roughly 90\%  scoring \texttt{5} (Figure \ref{fig:concept-result}). This is expected since here the original image is not input at all in producing the final image. \texttt{e2e-instruct} on the other hand makes no edit sometimes, with 40-60\% images being given a score of \texttt{1}. For countries like Brazil and US, this pipeline overwhelmingly paints the image with the flag or flag colors (\S\ref{app:e2e-instruct}), explaining the relatively lower number of \texttt{1}s.

\textbf{C1:} \texttt{semantic-equivalence} -- Here, we ask that if an edit is made (\textbf{C0}$<3$) is it a meaningful one? In Figure \ref{fig:concept-result}, we observe that \texttt{cap-edit} scores highest, while \texttt{cap-retrieve's} performance varies based on the country (lower for countries with low digital presence). 


\textbf{C2:} \texttt{spatial-layout} -- For \texttt{e2e-instruct} and \texttt{cap-retrieve}, we observe similar trends as those observed in \textbf{Q1)}. For \texttt{cap-edit}, while it scores mid to high on visual changes, it surprisingly maintains spatial layout, performing similar to \texttt{e2e-instruct}. This signifies that even though \texttt{cap-edit} makes visual edits, it does so while preserving spatial layout, helpful for audiovisual translation like in Doraemon, Inside Out and so on.

\textbf{C3:} \texttt{culture-concept} -- Each original image's cultural relevance score may be different to begin with. Hence, here we plot the delta in scores, relative to the original image. If $\mathrm{score_{edited} < score_{original}}$, we bucket it into $\mathrm{-\Delta}$ (\emph{negative change}); if $\mathrm{score_{edited} = score_{original}}$, we bucket it into \texttt{0} (\emph{no change}), and if $\mathrm{score_{edited} > score_{original}}$, we bucket it into $\mathrm{+\Delta}$ (\emph{positive change}). We observe that \texttt{cap-retrieve} performs best across all countries, followed by \texttt{cap-edit} and finally \texttt{e2e-instruct}. This indicates that while end-to-end image-editing models still have a long way to go in understanding cultural relevance, LLMs can take the responsibility of cultural translation and provide them with concrete instructions for editing or retrieval.

\textbf{C4:} \texttt{naturalness} -- \texttt{cap-retrieve} receives highest scores here since these are natural images retrieved from the internet. \texttt{cap-edit} receives a significant number of 4s, because it doesn't look as natural as retrieved images, but probably natural enough, as discussed in \S\ref{app:cap-edit}.

\textbf{C5:} \texttt{offensiveness} -- Almost no images are found to be offensive, which is encouraging.

\textbf{C1+C3:} \texttt{meaningful-edit} -- We plot counts of pipelines that score above \texttt{3} on \texttt{semantic- equivalence} and have a positive change in \texttt{culture-concept} score ($\mathrm{+\Delta}$). These images have been edited such that they increase cultural relevance while staying with bounds of the universal category, which is our end-goal for \emph{concept}. From Figure \ref{fig:concept-result}, we can see that performance of the best pipeline is as low as 5\% for countries like Nigeria, indicating that this task is far from solved. \\

\subsection{Application Dataset}
\label{app:appl}

\textbf{E1:} \texttt{education-task} and \textbf{S1:} \texttt{story-text} -- Our observations are similar to what we observe for \textbf{C1:} \texttt{semantic-equivalence} in \emph{concept}. The retrieval pipeline is especially noisy, given that the requirement of "equivalence" here is that the edited image must be able to teach the same concept (for education) or match the text of the story (for stories), harder than simply matching a category.

\textbf{E/S1+E/S2}: \texttt{meaningful-edit} -- Similar to \textbf{C1+C3}, the count of images that increase cultural relevance, while preserving meaning as required by the end-application, is very low. For countries like Portugal, no pipeline is able to translate any image successfully. For some other countries, the best pipeline is able to translate 10-15\% of total images.

\subsection{Quantitative Metrics}
For image-editing, these typically capture how closely the edited image matches -- (i) the original image; and (ii) the edit instruction. Following suit, we calculate two metrics:
\textbf{a)} \emph{image-similarity}: we embed the original image and each of the generated images using DiNO-ViT \cite{caron2021emerging} and measure their cosine similarity; and \textbf{b)} \emph{country-relevance}: we embed the text -- \texttt{This image is culturally relevant to \{COUNTRY\}}, and the edited images using CLIP \cite{radford2021learning} and calculate their cosine similarity. We present results for both metrics in Figures \ref{fig:target_source_sim} and \ref{fig:image_sim}. A discussion on correlation of these metrics with human evaluation is in \S\ref{app:quant_eval}.

We find that overall for \emph{image-similarity}, \texttt{e2e-instruct} scores highest, closely followed by \texttt{cap-edit}, while \texttt{cap-retrieve} lags behind, consistent with human ratings. For the \emph{country-relevance score}, we observe a similar trend as that for \textbf{C3:} \texttt{cultural-relevance}.


\begin{figure*}
    \centering
    \includegraphics[width=\linewidth]{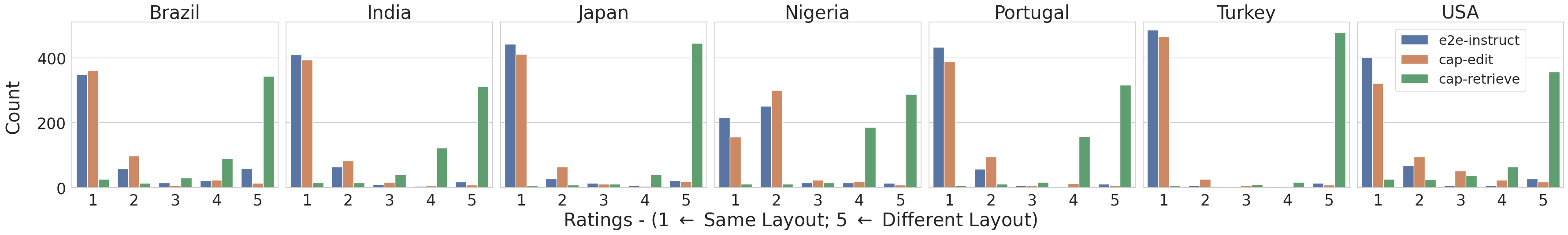}
    \caption{\textbf{Q3}: \texttt{spatial-layout}, capturing if the structure of the original image is maintained.}
    \label{fig:q3_part1}
\end{figure*}

\begin{figure*}
    \centering
    \includegraphics[width=\linewidth]{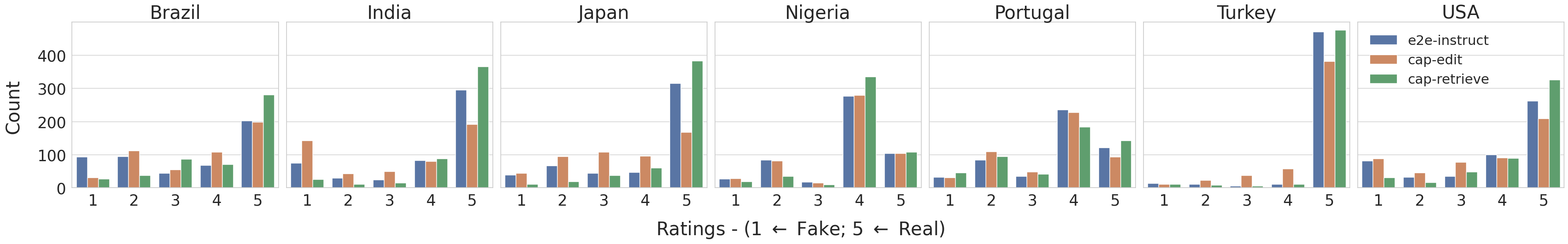}
    \caption{\textbf{Q5}: \texttt{naturalness} capturing the naturalness of the edited or retrieved image.}
    \label{fig:q5_part1}
\end{figure*}

\begin{figure*}
    \centering
    \includegraphics[width=\linewidth]{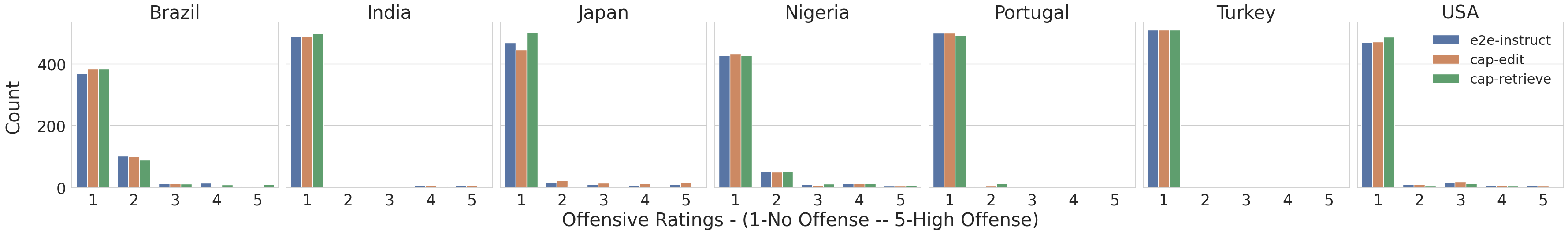}
     \caption{\textbf{Q6}: \texttt{offensiveness} capturing how offensive each pipeline is}
    \label{fig:q6_part1}
\end{figure*}

\onecolumn
\begin{tcolorbox}[colback=white,colframe=black,boxrule=0.5mm,arc=4mm,title=Prompts used for the pipelines described in Section 2.]
\ttfamily 
\textbf{InstructBLIP Prompt (Captioning)} \\
\underline{\smash{Concept Dataset}} \\
"A short image description:" \\
\underline{\smash{Application Dataset (Education)}} \\
"This image is from a math worksheet titled: TASK. Describe the image such that it talks about details relevant to the task of the worksheet. The output should be ONLY ONE sentence long." \\
\underline{\smash{Application Dataset (Stories)}} \\
"This image is from a storybook for children. Caption the image such that it describes details relevant to the story." \\

\textbf{GPT3.5 Prompt (LLM-editing)} \\
\underline{\smash{Concept Dataset}} \\
"Edit the input text, such that it is culturally relevant to \texttt{COUNTRY}. Keep the output text of a similar length as the input text. If it is already culturally relevant to \texttt{COUNTRY}, no need to make any edits. The output text must be in English only.\\Input: " \\
\underline{\smash{Application Dataset (Education)}} \\
"Edit the input text, such that it is culturally relevant to \texttt{COUNTRY}. The text describes an image in a math worksheet titled: TASK. Hence, make sure the edit preserves the intent of the task in the worksheet. Keep the output text to be of a similar length as the input text. If it is already culturally relevant to \texttt{COUNTRY}, there is no need to make any edits. The output text must be in English only.\\Input: " \\
\underline{\smash{Application Dataset (Stories)}} \\
"Edit the input text, such that it is culturally relevant to \texttt{COUNTRY}. The text describes an image in a storybook for children. Make sure the edit preserves the meaning of the story. Keep the output text to be of a similar length as the input text. If it is already culturally relevant to \texttt{COUNTRY}, there is no need to make any edits. The output text must be in English only.\\Input: " \\

\label{tab:prompts}
\end{tcolorbox}

\twocolumn

\end{document}